\documentclass[11pt]{article}
\usepackage{fullpage}

\usepackage{amsmath,amsfonts,bm}

\def\eqref#1{(\ref{#1})}

\def\1{\bm{1}}

\DeclareMathAlphabet{\mathsfit}{\encodingdefault}{\sfdefault}{m}{sl}
\SetMathAlphabet{\mathsfit}{bold}{\encodingdefault}{\sfdefault}{bx}{n}

\DeclareMathOperator*{\argmax}{arg\,max}
\DeclareMathOperator*{\argmin}{arg\,min}

\DeclareMathAlphabet\mathbfcal{OMS}{cmsy}{b}{n}

\newcommand{\red}{\color{red}}

\newcommand{\cmark}{\textcolor{green}{\ding{51}}}
\newcommand{\xmark}{\textcolor{red}{\ding{55}}}

\usepackage{multirow,wrapfig}
\usepackage{graphicx}
\usepackage{hyperref}
\usepackage{subfigure}
\usepackage{graphicx}
\usepackage{booktabs}
\usepackage{pifont}
\usepackage{todo}
\usepackage{algorithm,algpseudocode}
\usepackage{mathtools}
\usepackage{color}
\usepackage{adjustbox}
\usepackage[most]{tcolorbox}
\usepackage{wrapfig}
\usepackage{cite}
\usepackage[T1]{fontenc}
\usepackage{pifont}
\usepackage[para,online,flushleft]{threeparttable}
\usepackage{caption}
\usepackage{booktabs,pifont}
\usepackage[capitalize,noabbrev]{cleveref}
\newcommand{\bcc}[1]{\left\{{#1}\right\}}
\newcommand{\brr}[1]{\left({#1}\right)}
\newcommand{\bs}[1]{\left[{#1}\right]}

\usepackage{multirow}
\usepackage{amssymb}
\usepackage{amsfonts}
\usepackage{graphics}
\usepackage{graphicx}
\usepackage{setspace}
    
\newcommand{\lv}[1]{%
    \ifmmode
    \text{\textcolor{orange}{[LV: #1]}}
    \else
    \textcolor{orange}{[LV: #1]}
    \fi
}
\usepackage[utf8]{inputenc} 
\usepackage[T1]{fontenc}    
\usepackage{url}            
\usepackage{booktabs}       
\usepackage{amsfonts}       
\usepackage{nicefrac}       
\usepackage{microtype}      
\usepackage{xcolor}         
\usepackage{dsfont}
\usepackage{pifont}

\providecommand{\assumptionname}{Assumption}

\algnewcommand{\algorithmicforeach}{\textbf{for each}}
\algdef{SE}[FOR]{ForEach}{EndForEach}[1]
  {\algorithmicforeach\ #1\ \algorithmicdo}
  {\algorithmicend\ \algorithmicforeach}

\newcounter{protocol}
\makeatletter
\newenvironment{protocol}[1][htb]{%
  \let\c@algorithm\c@protocol
  \renewcommand{\ALG@name}{Protocol}
  \begin{algorithm}[#1]%
  }{\end{algorithm}
}

\begin{document}
\title{Aligning Large Language Models with Human Feedback: Mathematical Foundations and Algorithm Design}
\author{Siliang Zeng\textsuperscript{1}, Luca Viano\textsuperscript{2}, Chenliang Li\textsuperscript{3}, Jiaxiang Li\textsuperscript{1}, Volkan Cevher\textsuperscript{2},\\  Markus Wulfmeier\textsuperscript{4}, Stefano Ermon\textsuperscript{5}, Alfredo Garcia\textsuperscript{3}, Mingyi Hong\textsuperscript{1} \footnote{This work has been accepted for publication in IEEE Signal Processing Magazine. This arXiv version includes Volkan Cevher as a coauthor in recognition of his substantial contributions. The IEEE Signal Processing Magazine special issue limits individual authors to appearing on at most two articles within the same issue, which determined the authorship listing in the journal version.}\\[4pt]
\textsuperscript{1}University of Minnesota, \textsuperscript{2}EPFL, \textsuperscript{3}Texas A\&M University,\\ \textsuperscript{4}Google DeepMind, \textsuperscript{5}Stanford University\\
}
\maketitle

\begin{abstract}

This article provides an introduction to the mathematical foundations and algorithmic frameworks used to align Large Language Models (LLMs) with human intentions, preferences, and values. We discuss standard alignment techniques, such as fine-tuning (SFT), reinforcement learning with human feedback (RLHF), and direct preference optimization (DPO). We also explore the theoretical underpinnings of learning from human preferences, drawing connections to inverse reinforcement learning (IRL) and discrete choice models. We present state-of-the-art algorithms in a tutorial style, discuss their advantages and limitations, and offer insights into practical implementation. Our exposition is intended to serve as a comprehensive resource for researchers and practitioners, providing both a foundational understanding of alignment methodologies and a framework for developing more robust and scalable alignment techniques.

\end{abstract}

\section{Introduction}
\noindent {\bf Background.} {As large language models (LLMs) have taken the world by storm, it is clear that generative AI systems will soon become ubiquitous in our lives. LLMs have been applied beyond chatbots and personal assistants to tackle complex challenges, including video gaming \cite{berner2019dota} and autonomous control \cite{bellemare2020autonomous}. 
{After the successful introduction and demonstrated efficacy of alignment training in LLMs \cite{bai2022training,ouyang2022training}, the concept of {\em alignment} has been playing an increasingly important role in the design and training of AI systems.} Loosely speaking, alignment refers to the performance guarantee that the AI system will generate outcomes that are intended or preferred by the human user without undesirable side effects or behaviors such as deception or manipulation. 


More technically, the LLM alignment problem involves fine-tuning or adjusting a base model that was originally trained on extensive, diverse datasets. While such a base model is trained to learn core behaviors and general knowledge, it often falls short when applied to specific tasks or to meeting the nuanced preferences of a human user. To better {\it align} the model, additional feedback is gathered from human experts who evaluate and/or rank the model's performance within a specific task context. {As demonstrated in \cite{ouyang2022training,guo2025deepseek}, reinforcement learning (RL) holds significant promise for LLM alignment by leveraging additional feedback to enhance their performance.
} This feedback embodies expert judgments
regarding both the accuracy of the model's responses and any particular preferences they may have. The goal is to refine the model to better meet these specific requirements, ultimately enhancing its task relevance and user satisfaction.
Generally, it is observed that after the alignment process, LLMs can follow human instructions well, and can avoid providing toxic and non-preferable responses. Please refer to  Fig. \ref{schematic} as an illustration of the alignment process.

{\bf Contributions of This Work.} Despite extensive study of the alignment problem in recent literature, this research area remains young, and it is still evolving rapidly with many unresolved theoretical challenges.  
A primary contribution of this article is to introduce the signal processing community to the fundamentals of alignment, highlight state-of-the-art algorithms and approaches, and clarify the intersection 
\begin{wrapfigure}{r}{0.5\textwidth}
\centering 
\includegraphics[width=0.5\textwidth,trim={0.6cm 3.2cm 8.6cm 5.0cm},clip]{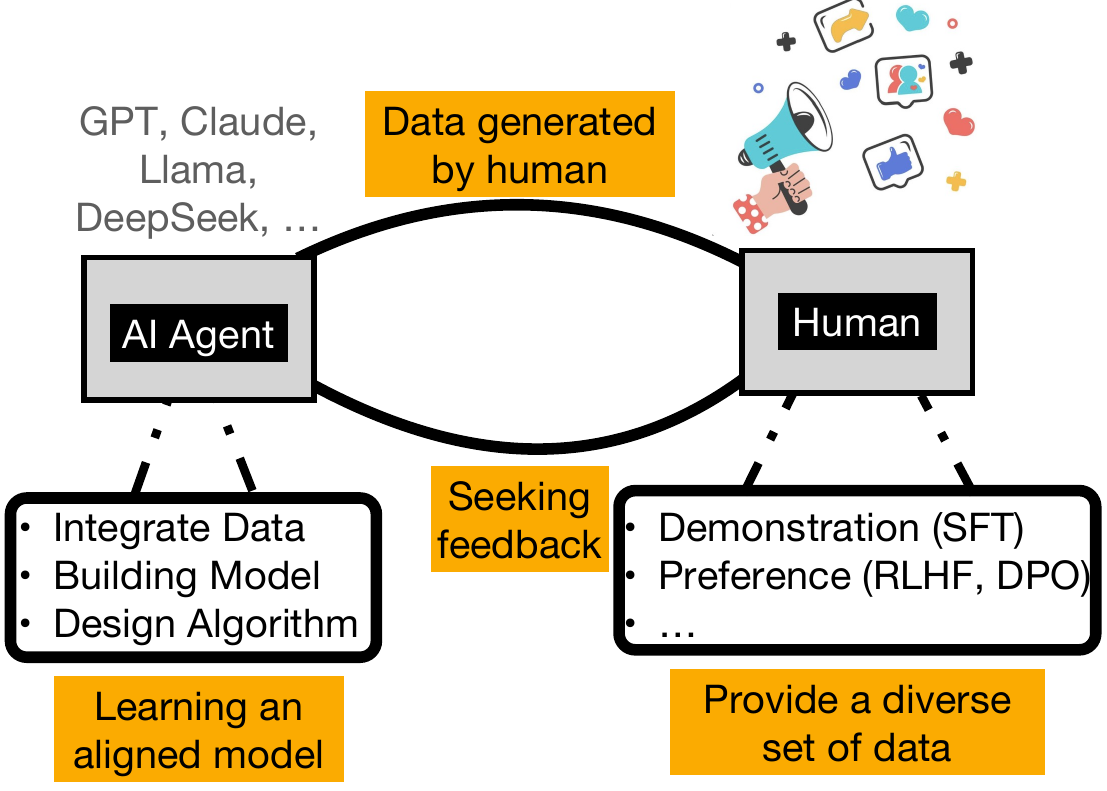}
\caption{\footnotesize An illustration of the alignment process.}\label{schematic}
\end{wrapfigure}
between alignment methodologies and various techniques relevant to signal processing (SP).
We believe that our article will foster interdisciplinary contributions and expand the role of SP in LLM alignment research. By understanding the alignment problem, SP researchers can leverage their expertise
in mathematical modeling, noise handling, inverse problem-solving, and optimization to address a number of key
challenges. For instance, similar to reconstructing
signals from observed data in SP, many alignment approaches involve solving inverse problems, such as learning the human preference models from the data. 
Therefore, we expect that introducing the alignment to SP professionals can foster innovation in
AI safety, inspire novel techniques, and accelerate the development of robust, generalizable alignment frameworks
that can be applied across diverse applications.

{\bf Organization of this work.} The article is organized as follows. In Sec. \ref{sec:prelim}, we provide the readers with an overview of the LLM training process. In Sec. \ref{sec:theoretical}, we discuss some classical notion and techniques of learning from human preference, which serves as the foundation for many alignment approaches to be discussed subsequently. In Sec. \ref{sec:sota}, we provide a detailed discussion of a number of state-of-the-art algorithms for different stages of LLM alignment. In Sec. \ref{sec:unified}, we provide readers with a framework that unify a few alignment algorithms discussed in the previous section under a single formulation. In Sec. \ref{sec:evals}, we provide some high-level discussion. 
In Sec. \ref{sec:conclusion}, we conclude with final remarks and open questions. 

\section{Preliminaries}\label{sec:prelim}
In this section, we provide a basic description of the LLM training process, while emphasizing the basic concepts relevant for alignment problems that we will describe in the sequel. 

\noindent{\bf LLM Representation.} Consider an LLM parameterized by parameters $\theta$ and denote the output probability by $\pi(y|x;\theta)$ where $x:=[x_1,\ldots,x_n]$ is the sequence of input prompts and $y:=[y_1,\ldots,y_m]$ is the sequence of output continuation, where $n$ and $m$ are their sequence length, correspondingly. In practice, each one of $y_i$ and $x_j$ is a token, which is one element of a token vocabulary. A token is a discrete unit that represents certain semantic information. For example, a word ``learning'' itself could be one token in the vocabulary, however more commonly two elements ``learn'' and ``-ing'' are both tokens where ``-ing'' represents the semantic information of tense. A tokenizer is commonly utilized to transform an input natural language sentence into sequence of tokens $x=[x_1,\ldots,x_n]$ and $y=[y_1,\ldots,y_m]$. Given sequences $x$ and $y$, we use $|x|$ and $|y|$ to denote the length of the sequence $x$ and $y$.

Typical LLM is an auto-regressive model, meaning that it predicts the output probability of the $y_j$ given all tokens in $x$ and $y_{<j}:=[y_1,\ldots,y_{j-1}]$ ($y_{<1}$ is null). More precisely, the 
\begin{wrapfigure}{r}{0.6\textwidth}
\centering
\includegraphics[width=0.6\textwidth,trim={1cm 10.5cm 0.2cm 5cm},clip]{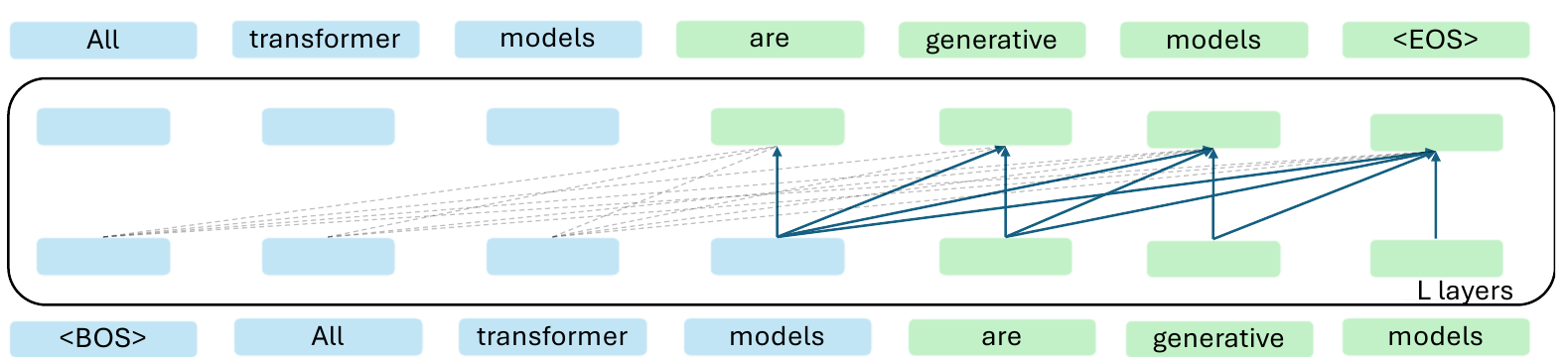}
\vspace{-0.4cm}
\caption{\footnotesize An illustration of Auto-regressive LLMs. Here <BOS> and <EOS> are two special tokens representing the start and the end of the sentence, respectively. The tokens at the bottom are the input token sequence, and those at the top are the output tokens. In this example, the token sequence ``<BOS> All transformer models '' is given to the LLM and ``are generative models <EOS>'' is the prediction. Each of the previous tokens will be used to predict the next token, corresponding to \eqref{eq:llm_probability}. 
}\label{fig:autoregrassive_model}
\end{wrapfigure}
probability of producing a sequence $y$ can be defined as:
\begin{equation}\label{eq:llm_probability}
\pi(y|x;\theta)=\prod_{j=1}^{m}\pi(y_j|x,y_{<j};\theta).
\end{equation}
The state-of-the-art LLMs are built upon the attention mechanism \cite{vaswani2017attention} and $\theta$ represents all the trainable parameters, including the attention matrices and weights of the feed-forward layers.

\noindent

\textbf{Model Architecture.} 
Before going to the details, we briefly discuss the standard architectural design adopted by modern large language models.
Most LLMs are built upon the Transformer~\cite{vaswani2017attention} architecture, which consists of a stack of self-attention and feed-forward (MLP)~\cite{rumelhart1986learning} layers interleaved with normalization~\cite{ba2016layer} and residual connections~\cite{szegedy2017inception}. 
Each input token $x_i$ is first embedded into a continuous representation and then processed through multiple attention heads, where self-attention~\cite{vaswani2017attention} computes contextualized token representations by weighting all previous tokens according to their learned relevance. 
Formally, each attention head computes $\text{Attn}(Q,K,V) = \text{softmax}(QK^\top / \sqrt{d_k})V$, where $Q,K,V$ are linear projections of the hidden states. 
The output of each attention block is then passed through the MLP layers or normalization layers such as LayerNorm~\cite{xu2019understanding}, forming one Transformer block. 
Stacking $L$ such blocks yields a deep auto-regressive model that predicts the next-token probability $\pi(y_j \mid x, y_{<j}; \theta)$ as defined in~(1). 
These components together define the function class $\pi(\cdot;\theta)$ over which subsequent alignment and optimization problems are formulated.

Next, we provide an overview of the different stages of LLM training. 
Before delving into the alignment-specific stages, it is worth noting that fine-tuning is widely used in LLM training beyond alignment tasks. 
Depending on the supervision signal, fine-tuning can take the form of supervised fine-tuning (SFT) or RL-based optimization. 
These approaches have been extensively applied to a variety of downstream objectives, including domain adaptation, instruction following~\cite{zhou2023instruction}, reasoning enhancement~\cite{wei2022chain,guo2025deepseek}, controllable generation~\cite{zhang2025aligning}, and agent training~\cite{qi2024webrl,team2025kimi}. For completeness, we briefly review both stages below.

\noindent {\bf The LLM pre-training.} The first stage of the training is called {\it pre-training}, where an LLM model is initialized randomly and trained over a vast corpus of documents, known as the pre-training dataset. As an example, Llama 3 model is pre-trained on more than 15T token of documents collected from all over the Internet, with the entire token vocabulary size (i.e., the total number of distinct tokens) 
of 128K~\cite{dubey2024llama}. 
The pre-training dataset is an unsupervised dataset, denoted as $\mathcal{D}_{\text{pre}}:=\{x^i\}_{i=1}^{N_{\text{pre}}}$, where each sample $x^i=[x^i_1,\ldots,x^i_{m_i}]$ is a sequence of $m_i$ tokens, and $m_i$ denotes the length of the sequence. $\mathcal{D}$ is a dataset without labels usually collected from the internet, such as Wikipedia articles; the learning task at the pre-training stage is typically 
formulated as the so-called {\it next-token prediction} task, which predicts the next token given the current and all previous tokens shown to the LLM, and can be mathematically formulated as a negative log-likelihood minimization problem:
\begin{equation}\label{eq:pretrain_loss}
    \min_{\theta} -\frac{1}{N_{\text{pre}}}\sum_{i=1}^{N_{\text{pre}}} \log \pi(x^i;\theta) = - \frac{1}{N_{\text{pre}}}\sum_{i=1}^{N_{\text{pre}}}\sum_{j=1}^{m_i} \log \pi(x_j^i|x_{<j}^i;\theta).
\end{equation}
This task is commonly understood as encouraging the LLM to memorize all the given texts and enabling the LLM for more fine-grained tasks. \eqref{eq:pretrain_loss} is further solved by first order optimizers such as SGD, RMSProp~\cite{hinton2012rmsprop} and AdamW~\cite{kingma2015adam,loshchilov2019decoupled}. In practice, AdamW is the default optimizer for pre-training, fine-tuning and aligning LLMs due to its efficiency and stability. We use AdamW to solve all the alignment protocols in this work unless otherwise mentioned.


\noindent {\bf The LLM Alignment.}  Once the pre-training is done, the next step is called {\it alignment} or {\it fine-tuning}. The fine-tuning process is a supervised process that aims at improving the instruction following capabilities of LLMs to better align with human behaviors and values. It usually consists of two main steps: the supervised fine-tuning (SFT) step and the reinforcement learning with human feedback (RLHF) step.

The SFT step utilizes a demonstration dataset $\mathcal{D}_{\text{demo}} :=\{(x^i,y^i)\}_{i=1}^{N_{\text{demo}}}$, where $x$ is the input prompt (such as, ``Where is the capital of the US?'') and $y$ is the expert response (such as, ``Washington, DC''). We assume that the demonstration continuations $y$ are collected from an \textit{expert} (typically a human expert, but sometimes it can also be a very large and well-aligned model such as GPT4 which produces high quality responses), thus we also denote $(x,y)\sim\mathcal{D}_{\rm demo}$ as $x\sim\rho, y\sim\pi^E(\cdot | x)$ as their population distributions, 
\begin{wrapfigure}{r}{0.4\textwidth}
\centering 
\includegraphics[width=0.4\textwidth,trim={0.5cm 7.2cm 11.8cm 3.6cm},clip]{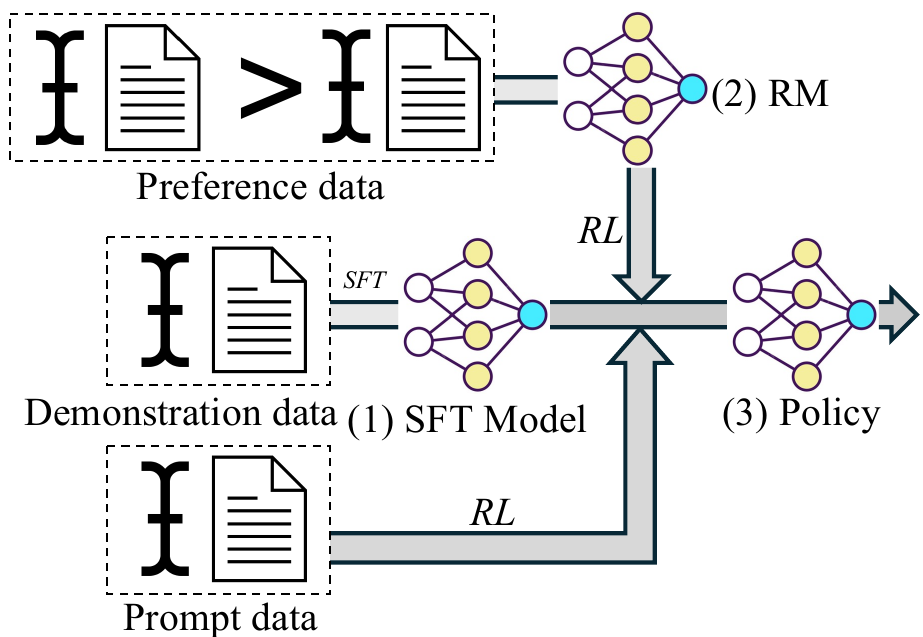}
\vspace{-0.7cm}
\caption{\footnotesize An illustration of the standard LLM Alignment process. The standard RLHF pipeline first trains a supervised model on demonstrations, then builds a reward model from preference data, and finally optimizes the policy with RL guided by the reward model.}\label{illustration}
\vspace{-0.1cm}
\end{wrapfigure}where $\rho$ is the distribution of the input prompts when collecting the data, and $\pi^E$ is the expert policy. 

Meanwhile the RLHF step first utilizes a {\it preference dataset} to learn from human preference about the quality of the answers to a particular question $x$. This dataset is denoted as $\mathcal{D}_{\text{pref}}:=\{(x,y_w\succ y_l)\}$, where $y_w$ is preferred (i.e., `\underline{w}ins') over $y_l$ (i.e., `\underline{l}oses') by a human labeler, and will be succinctly denoted as $(y_w\succ y_l)$ throughout the paper. We use the notation $x \sim \rho, (y_w\succ y_l)\sim\pi^{P} (\cdot | x)$ to denote the population distribution of the preference dataset, where $\pi^{P}$ is the preference distribution. The preference data is typically used to learn a parameterized {\it reward function} $r(x, y; \phi)$ where $\phi$ is the reward parameters. The reward model provides a score for a given prompt-response pair $(x,y)$.  Once such a reward function is learned, the RLHF step leverages this reward function, together with a {\it prompt dataset} $\mathcal{D}_{\rm prompt}:=\{x\}$, to evaluate and further improve the quality of the generated responses, through a reinforcement learning (RL) process. The whole process of RLHF is summarized and illustrated in Fig. \ref{illustration}. 


Below we provide an example of the demonstration and the preference data.


{\centering
\vspace{0.3cm}
\setlength\fboxrule{0.0pt}
\noindent\fcolorbox{black}[rgb]{0.95,0.95,0.95}
{\begin{minipage}{\columnwidth}
\noindent {\bf Example 1: Data of LLM Alignment.} Given a question prompt ``How should I respond to an email from my professor asking for a delayed assignment?'' The demonstration data includes a {\it preferred} answer ``You can politely acknowledge the delay and provide a reasonable explanation. For example: ‘Dear Professor, I apologize for the delay in submitting my assignment. I encountered unexpected challenges but will submit it by [new deadline]. Thank you for your understanding.'' Meanwhile, the preference dataset includes an additional  {\it non-preferred} answer, e.g.,  ``Just say you were busy and didn’t have time. Professors have to deal with it.'' Clearly, the preferred response aligns with human values of politeness and professionalism.
\end{minipage}
}}
\vspace{0.1cm}


We will provide a detailed account of popular methods to learn from these different kinds of datasets in the subsequent sections. In a high level, the SFT step can be used to directly teach LLM to imitate the expert behavior, while the RLHF step teaches LLM to distinguish the better behavior over the worse ones, in order to achieve a better generalization.




Based on the above discussion, we can now define the LLM alignment problem as follows: 

{\centering
\vspace{0.3cm}
\setlength\fboxrule{0.0pt}
\noindent\fcolorbox{black}[rgb]{0.95,0.95,0.95}{\begin{minipage}{1\columnwidth}
\noindent {\bf LLM alignment problem:} Given a diverse dataset representing human demonstrations and preferences, how can we effectively train an LLM to consistently generate outputs that align with human values, ethical principles, and intended use cases?
\end{minipage}}}    
\vspace{0.1cm}

\noindent {\bf The LLM Evaluation.} Before closing this introductory section, let us briefly discuss how the LLMs are evaluated after different stages of training. At the pre-training stage, the model learns to predict the next token probabilistically. The model is thus evaluated on how well it predicts the next token, and the common measure is the perplexity (PPL). 

Mathematically, given a sequence $x = [x_1,\ldots,x_m]$, perplexity is defined as the exponential of the average negative log-likelihood of the tokens:
\begin{equation*}
    \text{PPL}(x;\theta) := \exp\!\left( -\frac{1}{m}\sum_{j=1}^m \log \pi(x_j \mid x_{<j}; \theta) \right),
\end{equation*}
which assigns a single score to each sequence. A lower perplexity indicates that the model is more confident and accurate in predicting the next tokens.

In the fine-tuning stage, multiple metrics are typically used. After the SFT step, the metric is usually determined by the nature of the task. For classification tasks such as sentiment analysis (i.e., given a text of words, determine whether the sentiment of the text is positive or negative), 
 traditional metrics, such as accuracy, precision and recall are used; For translation tasks, BLEU (BiLingual Evaluation Understudy) score is the most popular metric, which measures how similar a machine-translated text is to a human-written reference translation. After the RLHF step, 
the aligned model could be evaluated by win-rates from human raters or some judge models  (typically takes the form of a specialized LLM), which calculate the favorability of the model response over other models/ground truth response. 

In the next section, we will begin our technical discussion by delving into classical techniques on learning from human feedback data. This will serve as mathematical foundation for various alignment techinques to be discuss subsequently.






\section{Theoretical Foundations of Learning from Human Feedback.}\label{sec:theoretical}

Given the background above, it is clear that the alignment problem is closely related to the problem of modeling human behaviors through the observed  choices. Indeed, in the context of LLM, the response $y$ in the demonstration data as well as the labels that distinguish $y_w$'s with $y_l$'s can both be viewed as human choices. It turns out that there is a vast literature about developing models for discrete choices, and these models were first used to describe discrimination between perceptual stimuli \cite{thurstone1927law} and choices for urban transportation modes \cite{McFadden1974}. In this section, we will provide a brief overview of this literature. 

\subsection{Classical models of discrete choice}\label{sub:discrete}
The choice models are used to describe observed choices (or actions) $a \in A$ made by a decision maker given a relevant state variable $s \in S$. A parallel to LLM alignment problem described in the previous section can be made by having the state $s$ correspond to the prompt $x$ 
and the action $a$ corresponds to the possible responses $y$'s to the prompt.

As in the previous section, we use the notation $(s,a) \succ (s,a')$ to indicate that in state $s$ action $a \in A$ is preferred over $a'\in A$. An assumed {\it structure} of preferences over action set $A$ is a relation $\succ$, which is \\
\noindent$\bullet$~{\em Complete}: $(s,a)\succ (s,a')$ or $a'\succ a$ for all $a,a' \in A$ and $s \in S$.

\noindent$\bullet$~{\em Transitive}: $(s,a)\succ (s,a')$ and $(s,a')\succ (s,a'')$ implies $(s,a)\succ (s,a'')$, for all $a,a',a'' \in A$, $s \in S$.

A structure of preferences can be modeled by a reward (or utility) function $r:S\times A \mapsto \mathbb{R}$, such that 
$(s,a) \succ (s,a')$ if and only if  $r(s,a) > r(s,a')$. This implies  when in state $s$, the decision maker will always prefer $a$ to $a'$. However in practice, one may observe a decision maker engages in {\it mixing}, i.e. implementing different actions when in the same state $s \in S$. In the context of LLM alignment, mixing corresponds to observing different responses to the same prompt.

There are several ways of modeling mixing in the literature of discrete choice:

\noindent {\bf Random Utility}: 
In this model, the decision maker perceives a random reward (or utility) $r(s,a) + \epsilon(a)$ (where  $\epsilon(a)$ is a random variable) given the pair $(s,a)$.  A pair $(s,a)$ is preferred to $(s,a')$ if:
 $$r(s,a) + \epsilon(a) > r(s,a') +\epsilon(a').$$
For a given state $s\in \mathcal{S}$ and a reward $r(\cdot)$, the probability of such preference can be modeled as follows:
 \begin{align*}
 P_r\Big((s,a) \succ (s,a')\Big) &:=\mbox{Prob}\Big(r(s,a) + \epsilon(a) > r(s,a')+\epsilon(a')\Big) = \mbox{Prob}\Big(\epsilon(a)-\epsilon(a') > r(s,a')-r(s,a)\Big).
 \end{align*}
 When the random variables $\{\epsilon(a):a\in A\}$ are assumed i.i.d and {standard Gumbel} we obtain the following logit model \cite{Luce59,bradley1952rank} {(also known as the Bradley-Terry-Luce model (BTL))}:
 \begin{align}
 P_r\Big((s,a) \succ (s,a')\Big)&=\frac{\exp r(s,a)}{\exp r(s,a)+\exp r(s,a')} =\sigma\Big(r(s,a)-r(s,a')\Big), \label{eq:random:utility}
 \end{align}
 where $\sigma(x):=\frac{1}{1+e^{-x}}$.
{In \cite{hamilton2023many}, the authors summarize the various motivations and use-cases of the BT model.}
We will see later that the logit model expressed above has been heavily used in the RLHF step to learn a reward function from preference data. 

\noindent {\bf Perturbed Utility}: Stochastic choice can also arise as the solution to expected reward maximization subject to an additive perturbation \cite{Machina1985} which may describe limited information
processing capacity \cite{Tishby2011} or aversion to ambiguity \cite{Maccheroni2006}. Assuming that the reward function $r(\cdot)$ is known, then the optimal stochastic choice can be modeled as the following utility maximization problem 
\begin{align}\label{eq:pi:star}
   \pi^*_r(\cdot|s):= \arg \max_{\pi \in \Delta^{|A|}}\{\mathbb{E}_{a \sim \pi}[r(s,a)]- \beta\times c(\pi)\},
\end{align}
where $\Delta^{|A|}\subset \mathbb{R}^{|A|}$ is the simplex; $c:\Delta^{|A|}\mapsto \mathbb{R}$ is a convex function; $\beta>0$ is a constant. For example, when $c$
is the (forward) Kullback-Leibler (KL) divergence with respect to a default policy $\pi^0$ defined as
\begin{align}\label{eq:kl}
 D_{\rm KL}(\pi \| \pi^0) := \sum_{a\in A}\pi(a)\log \frac{\pi(a)}{\pi^0(a)},
\end{align}
then the optimal stochastic choice can be expressed as follows:
\begin{align}
\pi^*_r(a|s)=\frac{\pi^0(a|s)\exp (\beta^{-1}r(s,a))}{\sum_{\tilde{a}\in A} \pi^0(\tilde{a}|s)\exp(\beta^{-1}r(s,\tilde{a}))}.
\label{eq:random:pertubed_utility}
\end{align}
It should be noted that in order to precisely calculate the optimal policy, one has to assume that all available actions in $A$ are known. Nevertheless, the perturbed utility model described here is instrumental for establishing a number of LLM alignment approaches, assuming that a reward model is available.


\subsection{The reward learning problem}
Based upon the observed preferences $\mathcal{D}_{\rm pref}:=\{(s,a)\succ (s,a')\}$ by an agent and the model \eqref{eq:random:utility}, the estimation problem consists of finding a reward function $r(s,a)$ that {\em rationalizes} the data, so  that observed preferences are {\it consistent} with a reward model. Specifically, the likelihood of $\mathcal{D}_{\rm pref}$ can be written as 
\begin{align}\label{eq:likelihood:rm}
    \ell_{\rm pref}(r)=\mathbb{E}_{ \mathcal{D}_{\rm pref}}\Big[\log P_r\Big((s,a) \succ (s,a')\Big)\Big].
\end{align}
In general, there is no unique reward function that maximizes likelihood $\ell_{{\rm pref}}(r)$. However, assuming the reward $r(s,a_0)$ for a reference action $a_0 \in A$ is known, it can be shown a unique reward function that maximizes likelihood function $\ell_{\rm pref}(r)$.

Alternatively, based upon the observed choices $\mathcal{D}_{\rm demo}:=\{(s,a)\}$ and assuming these choices are {\it consistent} with the model \eqref{eq:random:pertubed_utility}. Then the estimation problem consists of finding a reward function $r(s,a)$ that maximizes the likelihood of $\mathcal{D}_{\rm demo}$ defined as:
\begin{align}\label{eq:likelihood:sft}
\ell_{\rm demo}(r)=\mathbb{E}_{ \mathcal{D}_{\rm demo}}\Big[\log \pi^*_r(a|s)\Big].
\end{align}
As pointed out before, a limitation of this model is the assumption of complete information on the menu of available choices $A$.  
{When the reward is linearly parametrized, i.e. $r(s,a)=\phi(s,a)^{\top}\theta$ where $\phi(s,a),\theta \in \mathbb{R}^p$ and $\phi(s,a)$ is a vector of features, problem \eqref{eq:likelihood:sft} is the Lagrangian dual of the {\em maximum entropy} estimation problem (see Theorem 2 in \cite{zeng2022structural}):
\begin{align*}
\max_{\pi(\cdot|s) \in \Delta^{|A|}}& \quad \mathbb{E}_{s\sim \mathcal{D}_{\rm demo}}[\mathcal{H}(\pi(\cdot|s))], \quad {\rm s.t.} \quad \mathbb{E}_{s\sim \mathcal{D}_{\rm demo},a \sim \pi(\cdot|s) }[\phi(s,a)].
\end{align*}

}


\section{State-of-the-art Algorithms for LLM Alignment}\label{sec:sota}

In this section, we provide an overview of the mathematical formulation of the alignment problem, and a number of state-of-the-art algorithms. To facilitate discussion, this section is organized according to the \emph{structure} of the available data. 

\subsection{Learning from demonstrations data}
We begin with the discussion on  how  to model and learn from \emph{demonstration data} $\mathcal{D}_{\text{demo}}:=\{(x, y)\}$. Given the prompt $x$, the response $y$ is viewed as a {\it golden answer}. A natural approach to learning from such a dataset is supervised learning, leading to the supervised fine-tuning (SFT) paradigm. To this end, 
we formulate the following optimization problem over the LLM parameters class $\Theta$: 
\begin{equation}\label{eq:SFT}
\min _{\theta \in \Theta} \ \ell_{\mathrm{SFT}}(\theta):= - \mathbb{E}_{x\sim\rho, y\sim\pi^{\rm E}(\cdot | x)}\left[\log \pi\left(y | x;\theta\right)\right]
\end{equation} 
where $\pi^{\rm E}$ is the expert policy. It is easy to see the above problem shares the same optimal solutions as 
\begin{equation}
\min_{ \theta \in \Theta } ~\mathbb{E}_{x \sim \rho}[D_{\mathrm{KL}}(\pi^{\mathrm{E}}\left(\cdot | x\right)\|\pi\left(\cdot | x;\theta\right))],  \label{eq:SFT_eq}
\end{equation}
where $D_{KL}(\cdot)$ is the (forward) KL-divergence defined in \eqref{eq:kl}. 
The latter shows that SFT aims at imitating the demonstration dataset via minimizing the KL divergence.
Given a dataset $\mathcal{D}_{\rm demo}$ of expert demonstrations, we show the SFT method in Protocol~\ref{prot:SFT} where $ \widehat{\ell}_{\mathrm{SFT}}(\theta) $ is a finite sample version of $\ell_{\mathrm{SFT}}(\theta)$ in \eqref{eq:SFT}.

It is worth noting that the SFT stage described here is closely  
related to behavioral cloning \cite{pomerleau1988alvinn,ross2010efficient}, whose goal is to {\it mimic} the policy of an expert, we describe the details in the next paragraph.

\subsubsection{Behavioral cloning approaches} \label{sec:cloning}
Behavioural cloning (BC, see Protocol~\ref{prot:SFT}) it is an {\it offline} problem where the learner cannot collect new data using the current policy, but should learn relying exclusively on the expert dataset \cite{pomerleau1988alvinn}. 
This can be done via a reduction to supervised learning where the actions are the labels to be predicted and the states are the features.
Very recently, \cite{foster2024behavior} show that exact minimization of the loss $\widehat{\ell}_{\mathrm{SFT}}$ allows to output a policy which enjoys suboptimality guarantees compared to the expert. 
Their experiments and theoretical analysis also show that the guarantees and practical performance of the method do not deteriorate for long horizon problems. Unfortunately, exact minimization of the loss $\widehat{\ell}_{\mathrm{SFT}}$ is not possible for non-convex policy classes, and to our knowledge, there are currently no studies investigating how those optimization errors affect performance. 
In \cite{wulfmeier2024imitating}, the authors proposed offline inverse reinforcement learning (IRL)-based imitation leraning methods in the SFT setting to effectively utilize the sequential structure underlying autoregressive generation. Compared with standard SFT method, the IRL methods can extract both reward and policy from the demonstration dataset. Moreover, the authors find that offline IRL-based imitation retains diversity while
maximizing task performance. In this sense, IRL can also be regarded as a strong alternative on fixed SFT datasets even without online data generation. In Sec. \ref{sec:unified}, we will discuss the details of IRL in LLM alignment. It is also important to notice that in the alignment of LLM, the environment is simpler than what generally assumed in analyzing and empirically testing BC. Indeed, in language tasks LLMs are tree shaped and deterministic.


\begin{protocol}[t]
\caption{SFT (Supervised Fine Tuning) via Behavioral Cloning (BC) \label{prot:SFT}}
    \centering
  \begin{algorithmic}[1]
  \State The learner receives: 
(i)  A dataset $\mathcal{D}_{\text{demo}} = \{(x^i,y^i)\}^{N_{\mathrm{demo}}}_{i=1}$, 
(ii) A policy function parameters class $\Theta$.
  \State The learner computes the loss: $\widehat{\ell}_{\mathrm{SFT}}(\theta):= \mathbb{E}_{(x, y) \sim \mathcal{D}_{\rm demo}} \left[ - \log \pi\left(y^i | x^i;\theta\right)\right].$
  \State The learner outputs $\pi_{\theta^\star}$ with $\theta^\star = \argmin_{\theta\in\Theta} \widehat{\ell}_{\mathrm{SFT}}(\theta)$.
  \end{algorithmic}
\end{protocol}

\subsubsection{Self generation approaches}\label{sub:spin}
An alternative approach, dubbed SPIN (Self-Play fIne-tuNing), presented in \cite{chen2024self} casts the fine-tuning with demonstration problem as a two-player game.
On the one hand, the first player tries to generate answers as similar as possible as the observed ones in $\mathcal{D}_{\mathrm{demo}}$. On the other hand, the second player's aims at distinguish artificially generated answers from the answers in the demonstrations dataset $\mathcal{D}_{\mathrm{demo}}$.
The algorithm employs alternating updates between the first and second player. In particular, at iteration $t$, the first player observes the second players parameters $\theta_t$ and updates its weights via the following minimization over potentially time-varying parameters classes $\{\mathcal{F}_t\}$ as
\begin{equation}
f_{t+1} = \argmin_{f\in \mathcal{F}_t} \mathbb{E}_{x, y \sim \mathcal{D}_{\mathrm{demo}}} \mathbb{E}_{y' \sim \pi(\cdot|x;\theta_t)} \bs{\ell ( f(x, y) - f(x, y')) }, \label{eq:SPIN_f_update}
\end{equation}
where $\ell(\cdot)$ is a monotonically decreasing, non negative, smooth and convex function.
A common choice in practice satisfying the above properties is  the logistic loss function $\ell(x) = \log \big( 1 + \exp(-x) \big)$.
All these properties are imposed to make the optimization problem easier to solve via first order methods.
At this point, the second player can update its weights exploiting knowledge of $f_{t+1}$. This is done by solving the following problem:
$
\theta_{t+1} = \argmin_{\theta \in \Theta} - \mathbb{E}_{x \sim \rho, y \sim \pi(\cdot|x;\theta_t)}\bs{f_{t+1}(x,y) + \beta D_{\mathrm{KL}}(\pi(\cdot|x;\theta) || \pi(\cdot|x; \theta_t)) },$
which is an application of \Cref{eq:pi:star} for $c(\pi) = - D_{\mathrm{KL}}(\pi(\cdot|x) || \pi(\cdot|x;\theta_t)  )$.
The analytical solution of the above problem is surprisingly easy to obtain (analogously to \eqref{eq:random:pertubed_utility}): 
$
\pi(y|x;\theta_{t+1}) \propto \pi(y|x;\theta_{t}) \exp \brr{\frac{f_{t+1}(x,y)}{\beta}},
$
if the class $\Theta$ is expressive enough compared to the class $\mathcal{F}_{t+1}$. 
\noindent To ensure that this is the case, in \cite{chen2024self} the class $\mathcal{F}_{t+1}$ is chosen to be coupled with $\Theta$ as follows, $\mathcal{F}_{t+1} = \left\{ \beta \log \brr{\frac{\pi(y|x;\theta_{t+1})}{\pi(y|x;\theta_{t})}}\right\}$. Another benefit of this choice is that the two updates can now be expressed as single update over the weight class $\Theta$.
Indeed, \eqref{eq:SPIN_f_update} is at this point equivalent to 
\begin{align}\label{eq:spin}
\theta_{t+1} = \argmin_{\theta\in \Theta} \widehat{\ell}_{\mathrm{SPIN}}(\theta) =  \mathbb{E}_{(x, y) \sim \mathcal{D}_{\mathrm{demo}}} \mathbb{E}_{y' \sim \pi(\cdot|x;\theta_t)} \bs{\ell \left( \beta \log \brr{\frac{\pi(y|x;\theta)}{\pi(y|x;\theta_{t})}} - \beta \log \brr{\frac{\pi(y'|x;\theta)}{\pi(y'|x;\theta_{t})}}\right) } .
\end{align}
\begin{wrapfigure}{r}{0.5\textwidth}
    \centering
    \includegraphics[clip,width=0.75\linewidth]{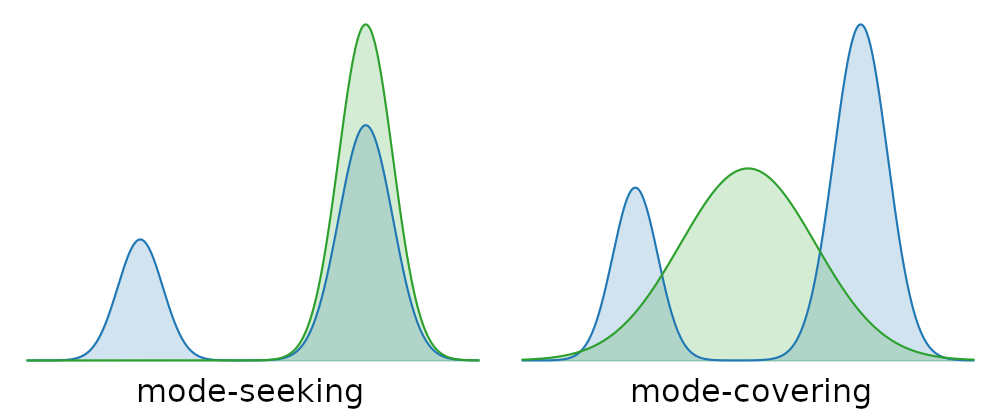}
    \caption{ \label{fig:mode_seeking_vs_covering} \footnotesize True distribution in blue and estimated distribution in green. When the estimated distribution has a smaller support than the true one, as in the left plot, we say that it has a mode seeking. Vice versa, the estimated distribution has a mode covering behavior if its support is larger, as in  the right plot. Image from \url{https://sander.ai/2020/03/24/audio-generation.html}.  }
    \label{fig:distribution_estimation}
\end{wrapfigure}
The practical algorithm can be obtained via a stochastic estimator of this loss constructed with the dataset $\mathcal{D}_{\mathrm{demo}}$, as described in Protocol~\ref{prot:SPIN}.
We note that SPIN shares similarity with Direct Preference Optimization (DPO), to be described in Section~\ref{sec:DPO}, however there are also important differences that we will highlight in Section~\ref{sec:unified}.

In SPIN, the adversarial imitation procedure leverages self-generated data to recover a demonstration-level policy (see Equation~\eqref{eq:SPIN_f_update}). Notably, both adversarial imitation frameworks like TextGAIL \cite{wu2021textgail} (see Protocol \ref{prot:GAIL}) and online IRL approaches \cite{zeng2025demonstrations} similarly incorporate self-generated rollouts into their training pipelines to enable effective learning from demonstrations. We discuss these details of IRL approaches and their connections with other baselines in Section \ref{sec:unified}.


\begin{protocol}[t]
\caption{SFT (Supervised Fine Tuning) via SPIN (Self-Play fIne tuNing) \label{prot:SPIN}}
    \centering
  \begin{algorithmic}[1]
  \State The learner receives: 
(i) A dataset $\mathcal{D}_{\text{demo}} = \{(x^i,y^i)\}^{N_{\mathrm{demo}}}_{i=1}$, 
(ii)  A policy function parameters class $\Theta$.
    \For{$t = 1, \dots, T$}
  \State The learner samples $y^{',i} \sim \pi(\cdot|x^i; \theta_t)$ and computes the loss: $$\widehat{\ell}_{\mathrm{SPIN}}(\theta) := \mathbb{E}_{(x, y) \sim \mathcal{D}_{\mathrm{demo}},\; y^{\prime} \sim \pi(\cdot \mid x; \theta_t)} \left[ \ell \left( \beta \log \left( \frac{\pi(y \mid x; \theta)}{\pi(y \mid x; \theta_t)} \right) - \beta \log \left( \frac{\pi(y^{\prime} \mid x; \theta)}{\pi(y^{\prime} \mid x; \theta_t)} \right) \right) \right].$$
  \State $\theta_{t+1} := \argmin_{\theta\in\Theta} \widehat{\ell}_{\mathrm{SPIN}}(\theta)$
  \EndFor
  \State The learner outputs $\pi_{\theta^\star}$ 
  \end{algorithmic}
\end{protocol}

\begin{protocol}[t]
\caption{GAIL (Generative Adversarial Imitation Learning) \cite{wu2021textgail} \label{prot:GAIL}}
    \centering
  \begin{algorithmic}[1]
  \State The learner receives: 
(i) A dataset $\mathcal{D}_{\text{demo}} = \{(x^i,y^i)\}^{N_{\mathrm{demo}}}_{i=1}$, 
(ii)  A policy function parameter class $\Theta$, (iii) A discriminator parameter class $\Phi$.
\State The learner optimizes the following problem: $$\min_{\theta} \max_{\phi} \widehat{\ell}_{\mathrm{GAIL}}(\theta, \phi) := \mathbb{E}_{(x, y) \sim \mathcal{D}_{\mathrm{demo}},\; y^{\prime} \sim \pi(\cdot \mid x; \theta)} \Big[ \log(1 - D_{\phi}(x, y)) + \log(D(x, y^\prime)) + \lambda \log \pi_{\theta}(y^\prime|x) \Big].$$
\State The learner outputs $\pi_{\theta^\star}$ 
\end{algorithmic}
\end{protocol}

\subsubsection{ Information theoretic approaches}
Recent works \cite{sun2024inverse,cundy2023sequencematch} considered matching general information theoretic divergences between answers distribution.

As mentioned, minimizing the forward KL divergence between demonstrated and learner answer distribution is equivalent to the SFT objective in \eqref{eq:SFT_eq}.
As commonly noticed, the forward KL minimization outputs a mode covering distribution. That is, the learned distribution has support which contains the support of the demonstrations distribution. 

 However, a mode seeking behavior can be more desirable if one wants to perfectly imitate some aspects of the demonstrations while discarding others. We refer the reader to \Cref{fig:mode_seeking_vs_covering} for a visual illustration. 
To this end, \cite{sun2024inverse} suggested to minimize the reversed KL divergence $\min_{ \theta \in \Theta } ~\mathbb{E}_{x \sim \rho}[D_{\mathrm{KL}}( \pi\left(\cdot | x;\theta\right)\| \pi^{\mathrm{E}}\left(\cdot | x\right) )]$.
This method however requires additional prompts and answers generated by the learner policy.
As a last example of divergence minimization, \cite{wulfmeier2015maximum} considers $\chi^2$ divergence minimization.
\subsection{Learning from preferences} \label{sub:preference:prompts}

Learning from preferences is usually performed in two steps: (i) reward learning using the preference dataset (ii) based on the reward model learned in the first step, further fine-tune the LLM by conducting reinforcement learning using a prompt dataset. Note that these two steps are closely related to the two types of stochastic choice models discussed in Section \ref{sub:discrete}.
\begin{protocol}[t]
\caption{RLHF (Reinforcement Learning from Human Feedback) \label{prot:RLHF}}
    \centering
  \begin{algorithmic}[1]
  \State The learner receives: 
(i)  A preference dataset $\mathcal{D}_{\rm pref} = \{(x^i,y^i_w, y^i_\ell)\}^{N_{\rm pref}}_{i=1}$,
(ii) A prompt dataset $\mathcal{D}_{\mathrm{prompts}} = \{ x^i\}^{N_{\mathrm{prompts}}}_{i=1}$,
(iii) A reward function parameter class $\Phi$,
(iv) A policy function parameter class $\Theta$.
\State The learner estimates the reward as $r\brr{\cdot,\cdot;\phi^\star}$ with 
\begin{equation*}
    \phi^\star = \argmin_{\phi\in\Phi}\ \widehat{\ell}_{\rm RM}(\phi)
    := - \mathbb{E}_{(x, y_w, y_l) \sim \mathcal{D}_{\rm pref}} \Big[ \log \Big( \sigma \big( r(x^i, y^i_w; \phi) - r(x^i, y^i_l;\phi) \big) \Big) \Big].
\end{equation*}
  
  \State The learner finds $\theta^\star = \argmin_{\theta\in\Theta} \ell_{\mathrm{RL}}(\theta)$, using RL trainer.
  \State The learner outputs $\pi_{\theta^\star}$.
  \end{algorithmic}
\end{protocol}

\begin{protocol}[t]
\caption{PPO (Proximal Policy Optimization) \cite{schulman2017proximal}\label{prot:PPO}}
    \centering
  \begin{algorithmic}[1]
  \State Receive policy class parameters $\Theta$. Initialize $\theta_1$.
    \For{$t = 1, \dots, T$}
    \State Sample $y \sim \pi(\cdot|x;\theta_t)$ and $x \in \mathcal{D}_{\mathrm{prompts}}$.
    
    \State Update the value function
    \State Compute advantages $\{ \hat{A}_{h} \}_{h = 1}^{|y|}$ according to generalized advantage estimation at the token level
    \State The learner computes \begin{align}
    \widehat{\ell}_{\rm PPO}(\theta) &= -\mathbb{E}_{x \sim \mathcal{D}_{\mathrm{prompts}}, y \sim \pi(\cdot|x; \theta_t)} \Bigg[ \frac{1}{|y|} \sum_{h=1}^{|y|} \Bigg\{ \min \Bigg[
        \frac{\pi(y_{h} | x, y_{<h}; \theta)}{\pi(y_{h} | x, y_{<h}; \theta_{t})} \hat{A}_{h}, \nonumber \\
        & \quad \quad \quad \quad \text{clip} \Bigg(
        \frac{\pi(y_{h} | x, y_{<h}; \theta)}{\pi(y_{h} | x, y_{<h}; \theta_{t})}, 
        1 - \epsilon, 1 + \epsilon 
        \Bigg) \hat{A}_{h}
        \Bigg]
        \Bigg\} \Bigg] \nonumber
\end{align}
    \State $\theta_{t+1} = \argmin_{\theta\in\Theta} \ell_{\mathrm{PPO}}(\theta) $ 
    \EndFor
  \State The learner outputs $\pi_{\theta^\star}$.
  \end{algorithmic}
\end{protocol}

\begin{protocol}[t]
\caption{REINFORCE \cite{williams1992simple} \label{prot:REINFORCE}}
    \centering
  \begin{algorithmic}[1]
  \State Receive policy class parameters $\Theta$. Initialize $\theta_1$, step size $\eta$
    \For{$t = 1, \dots, T$}
    \State Sample outputs $y \sim \pi(\cdot|x; \theta_t)$ and $x \in \mathcal{D}_{\mathrm{prompts}}$.
    \State The learner computes the gradient estimator \begin{align}
    \nabla \widehat{\ell}_{\rm REINFORCE}(\theta_t) &= -\mathbb{E}_{x \sim \mathcal{D}_{\mathrm{prompts}}, y \sim \pi(\cdot|x; \theta_t)} \big[ r(x, y) \nabla_{\theta} \log \pi(y | x; \theta_t) \big] \nonumber
\end{align}
    \State $\theta_{t+1} = \theta_t - \eta \nabla \widehat{\ell}_{\rm REINFORCE}(\theta_t)$
    \EndFor
  \State The learner outputs $\pi_{\theta^\star}$.
  \end{algorithmic}
\end{protocol}
\paragraph{Reward learning}
To find an appropriate reward model, RLHF (see e.g., ~\cite{christiano2017deep}) leverages a set of {\it preference dataset} $\mathcal{D}_{\rm pref}:=\{(x,y_w,y_l)\}$, where each data contains a pair of output $y_w,y_l$. The output $y_w$ is {preferred} over $y_l$ by human labeler (denoted as $y_w\succ y_l$). 
Leveraging the the Bradley-Terry model (BT) discussed in \eqref{eq:random:utility} and the likelihood of the preference data \eqref{eq:likelihood:rm}, 
one could formulate the following problem to find the reward model (parameterized by $\phi$): 
\begin{equation}\label{eq:RLHF_loss}
    \phi^\star = \argmin_{\phi \in \Phi}\ \ell_{\rm RM}(\phi)
    :=- \mathbb{E}_{x \sim \rho, (y_w\succ y_l)\sim\pi^{P}} \Big[ \log \Big( \sigma \big( r(x, y_w; \phi) - r(x, y_l;\phi) \big) \Big) \Big]. 
\end{equation}
where $(y_w\succ y_l)\sim\pi^{P}$ denotes sampling from the distribution of the preference dataset.
Protocol~\ref{prot:RLHF} describes the procedure where $r(\cdot; \phi)$ is learned via minimization of the empirical loss $\widehat{\ell}_{\mathrm{RM}}$. In the empirical loss $\widehat{\ell}_{\mathrm{RM}}$, we replace $(y_w\succ y_l)\sim\pi^{P}$ by sampling from a finite preference datset $(y_w\succ y_l)\sim \mathcal{D}_{\rm prefs}$.
\paragraph{Policy Learning}
Once a reward model $r(x,y;\phi^\star)$ is learned, the LLM can be fine tuned on the top of SFT by leveraging the perturbed utility model \eqref{eq:pi:star}, which results in the following problem:
\begin{equation}\label{eq:RLFT}
\min_{\theta} \ \ell_{\mathrm{RL}}(\theta):= -\mathbb{E}_{x \sim \rho, y \sim \pi(\cdot|x;\theta)}\left[r(x, y;\phi)\right]  + \beta\mathbb{E}_{x \sim \rho}[D_{\mathrm{KL}}(\pi\left(\cdot | x;\theta\right)\|\pi_{\mathrm{ref}}\left(\cdot | x\right))],
\end{equation}
where $\pi_{\mathrm{ref}}$ is a reference model such as the SFT-ed model, 
and $\beta$ is the regularization coefficient of the KL regularizer. Here, the optimal policy in \eqref{eq:RLFT} under the reward parameter $\phi$ has the following closed-form expression:
\begin{equation}
\pi_{r_\phi}^\star(y|x) = \frac{\pi_{\mathrm{ref}}(y|x)\exp(\beta^{-1} r(x,y;\phi))}{Z(x,\phi,\beta)},\label{eq:rlhf_optimal_policy}
\end{equation}
where we have $Z(x,\phi,\beta) := \sum_{y'} \pi_{\mathrm{ref}}(y'|x)\exp(\beta^{-1} r(x,y';\phi))$. 
Note that both the expected reward and the KL in \eqref{eq:RLFT} are {\it not} computable given the large cardinality of $\mathcal{X}$ and $\mathcal{Y}$,
so \eqref{eq:RLFT} is usually solved approximately by minimizing an empirical loss, using policy optimization techniques such as REINFORCE \cite{williams1992simple}, RLOO \cite{ahmadian2024back}, PPO \cite{schulman2017proximal} and GRPO \cite{shao2024deepseekmath}; see Protocol~\ref{prot:RLHF} - \ref{prot:GRPO}.

\emph{Pros and Cons: PPO or REINFORCE?} PPO was proposed as an improvement over the REINFORCE algorithm in environment where the gradient estimates generated by REINFORCE suffer from high variance. As a result, PPO enforces conservative policy updates to counterbalance the high variance effect. However, as argued in \cite{ahmadian2024back}, for the LLM fine-tuning task the gradient variance is not as high to justify the PPO conservative mechanism. 
Since both options are used across the community, we describe both algorithms in Protocols~\ref{prot:PPO} and~\ref{prot:REINFORCE} respectively. As a remark, PPO requires a critic model to estimate the advantage function which can be used to compute the value function score $ \{V(x, y_{<h})\}_{h = 1}^{|y|} $ at each token level and compute the token-level advantage estimation. Consequently, training the critic model for token-level advantage estimation will also incur additional memory overhead.
REINFORCE does not require training a critic, therefore it is more memory efficient than PPO. 
RLOO \cite{ahmadian2024back} is a modern REINFORCE modification that retains the advantage of not keeping a critic in memory and achieves a variance reduction effect subtracting the average of the other rewards (see Protocol \ref{prot:RLOO}). It can be shown that the RLOO estimator in an unbiased estimator of the advantage.
Recently, a new policy optimization method (GRPO) which does not require an estimated reward model has been introduced and has been proven successful in reasoning applications (see Protocol \ref{prot:GRPO} and Section~\ref{sub:verifier}). 

\begin{protocol}[t]
\caption{RLOO (REINFORCE Leave-One-Out) \cite{ahmadian2024back} \label{prot:RLOO}}
    \centering
  \begin{algorithmic}[1]
  \State Receive policy class parameters $\Theta$ and reward model parameters $\phi^\star$. Initialize $\theta_1$, step size $\eta$
        \For{$t = 1, \dots, T$}
    \State Sample $G$ outputs $\{y^{j}\}_{j=1}^{G} \sim \pi(\cdot|x; \theta_t)$ and $x \in \mathcal{D}_{\mathrm{prompts}}$.
    \State The learner computes rewards $\{r^j\}_{j=1,...,G}$ for each output $y^{j}$.
    \State The learner computes the estimated advantage function \begin{equation}
\hat{A}_{j} = r^j - \frac{1}{G - 1} \sum_{i \neq j} r^i. \nonumber
\end{equation}
    \State The learner computes \begin{align}
    \nabla \widehat{\ell}_{\rm RLOO}(\theta_t) &= -\mathbb{E}_{x \sim \mathcal{D}_{\mathrm{prompts}}, \{y^j\}_{j=1}^G \sim \pi(\cdot|x; \theta_t)} \Bigg[ \frac{1}{G} \sum_{j=1}^{G} \hat{A}^j \nabla_{\theta} \log \pi(y^j | x; \theta_t) \Bigg] \nonumber
\end{align}
    \State $\theta_{t+1} = \theta_t - \eta \nabla \ell_{\mathrm{RLOO}}(\theta_t)$
    \EndFor
  \State The learner outputs $\pi_{\theta^\star}$.
  \end{algorithmic}
\end{protocol}

\emph{Pros and Cons: Learning from demonstrations vs learning from preferences.}  It is widely observed that the models learned {\it only} using SFT stage do not perform as well as those that are trained via episodically learning the policy \eqref{eq:RLFT} and learning the reward \eqref{eq:RLHF_loss}  on top of the SFT models \cite{ouyang2022training}.
This is partly because the reward model and the RL algorithms together guide the LLMs to explore high-quality responses that can go beyond those defined by the SFT data, thus improving their generalization capabilities \cite{xu2020error}. 
Moreover, obtaining the SFT dataset is typically more expensive than obtaining preference data, therefore standard alignment process leverages {\it both} the SFT and the preference data.

\begin{protocol}[t]
\caption{GRPO (Group Relative Policy Optimization) \cite{shao2024deepseekmath}\label{prot:GRPO}}
    \centering
  \begin{algorithmic}[1]
  \State Receive policy class parameters $\Theta$, and reward model parameters $\phi^\star$. Initialize $\theta_1$.
    \For{$t = 1, \dots, T$}
    \State Sample $G$ outputs $\{y^{j}\}_{j=1}^{G} \sim \pi(\cdot|x; \theta_t)$ and $x \in \mathcal{D}_{\mathrm{prompts}}$.
    \State The learner computes rewards $\{r^j\}_{j=1,...,G}$ for each output $y^{j}$.
    \State The learner computes the estimated advantage function \begin{equation}
\hat{A}^{j} = \frac{r^j - \operatorname*{mean}(\{r^j\}_{j=1}^G)}{\operatorname*{std}(\{r^j\}_{j=1}^G)}, \quad j = 1, \cdots, G \nonumber
\end{equation}
    \State The learner computes \begin{align}
    \ell_{\rm GRPO}(\theta) &= -\mathbb{E}_{x \sim \mathcal{D}_{\mathrm{prompts}}, \{y^j\}_{j=1}^G \sim \pi(\cdot|x; \theta_t)} \Bigg[ \frac{1}{G} \sum_{j=1}^{G} \frac{1}{|y^j|} \sum_{h=1}^{|y^j|} \Bigg\{ \min \Bigg[
        \frac{\pi(y^{j}_{h} | x, y^{j}_{<h}; \theta)}{\pi(y^{j}_{h} | x, y^{j}_{<h}; \theta_{t})} \hat{A}_{j}, \nonumber \\
        & \quad \quad \quad \quad \text{clip} \Bigg(
        \frac{\pi(y^{j}_{h} | x, y^{j}_{<h}; \theta)}{\pi(y^{j}_{h} | x, y^{j}_{<h}; \theta_{t})}, 
        1 - \epsilon, 1 + \epsilon 
        \Bigg) \hat{A}_{j}
        \Bigg] - \beta \mathbb{D}_{\text{KL}}[\pi_{\theta} \| \pi_{\text{ref}}] 
        \Bigg\} \Bigg] \nonumber
\end{align}
    \State $\theta_{t+1} = \argmin_{\theta\in\Theta} \ell_{\mathrm{GRPO}}(\theta)$
    \EndFor
  \State The learner outputs $\pi_{\theta^\star}$.
  \end{algorithmic}
\end{protocol}


\subsection{Learning from preferences without online generations: Direct Preference Optimization}
\label{sec:DPO}
\begin{figure}{r}
    \centering
    \includegraphics[width=0.45\textwidth]{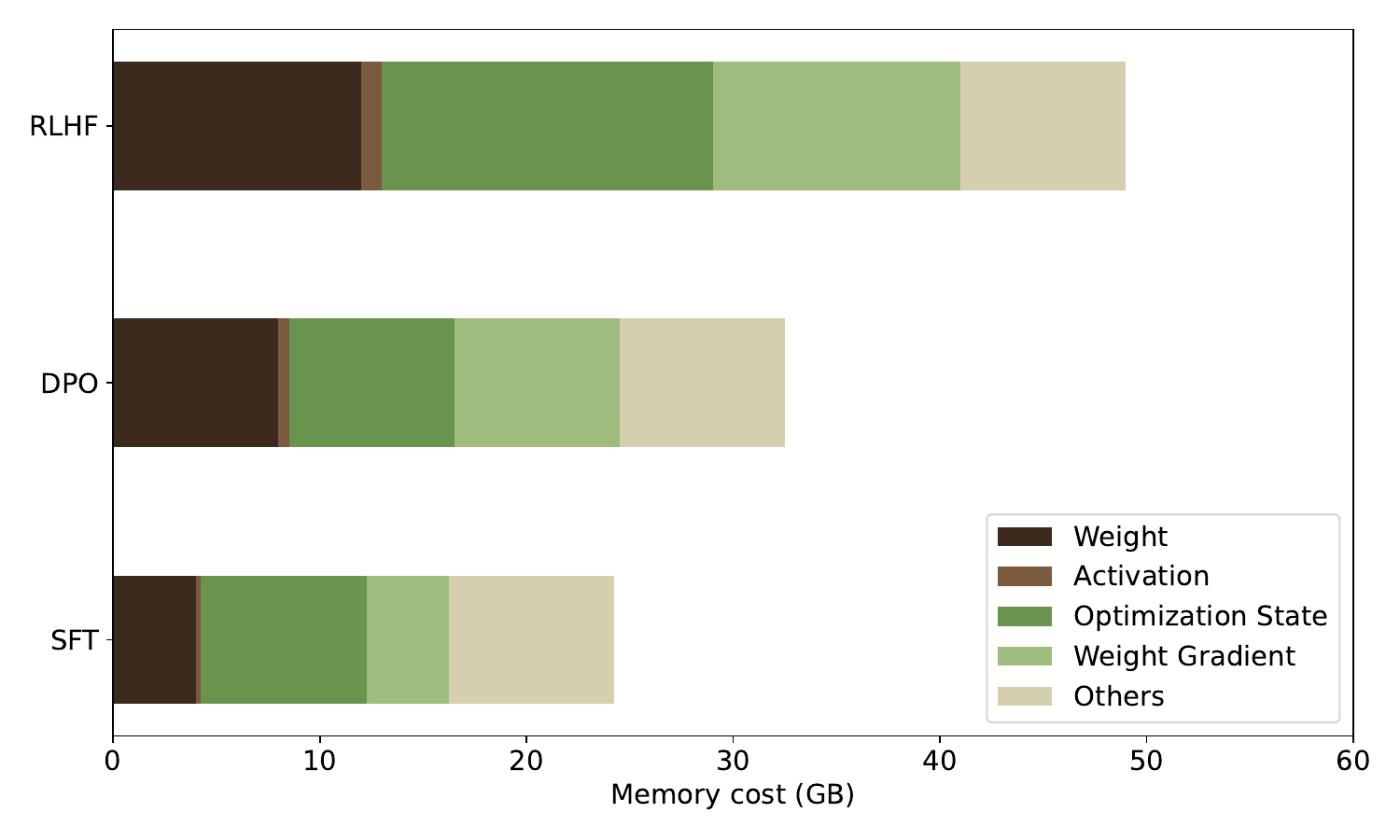}
    \caption{\footnotesize Estimated memory consumption of running the \texttt{Pythia-1B} model on \texttt{TL;DR} dataset with batch-size 2 on a single device}
    \label{fig:memory}
\end{figure}
Direct Preference Optimization  (DPO) \cite{rafailov2024direct} bypasses the need to perform the RL, by  noticing that minimizing $
\ell_{\mathrm{RL}}(\cdot)$ admits an analytical solution if we optimize over the whole Marvov stationary policy space denoted by $\Pi$.
In fact, notice that from our discussion on stochastic choice model \eqref{eq:random:pertubed_utility}, for a fixed $r(\cdot,\cdot; \phi)$, the solution to the policy optimization problem 
$
\pi_{\phi}^\star(\cdot, x) = \argmin_{\pi \in \Pi} - \mathbb{E}_{y \sim \pi(\cdot|x)}\bs{r(x,y;\phi)} + \beta D_{\mathrm{KL}}(\pi(\cdot|x) || \pi_{\mathrm{ref}}(\cdot|x))
$
is defined in \eqref{eq:rlhf_optimal_policy}.
This implies that for all $y$, it holds that $
\log (\pi_\phi^\star(y|x)) = \log ( \pi_{\mathrm{ref}}(y|x)) + \beta^{-1} r(x,y;\phi) - \log (Z(x,\phi,\beta)).$
Such a quantity cannot be computed in closed form because computing $Z(x,\phi,\beta)$ is intractable.
However, for two possible answers $y$, $y'$ to the {\it same} question $x$, it holds that
$\log (\pi_\phi^\star(y|x)) - \log (\pi_\phi^\star(y'|x)) = \log ( \pi_{\mathrm{ref}}(y|x)) - \log ( \pi_{\mathrm{ref}}(y'|x)) + \beta^{-1} r(x,y;\phi) - \beta^{-1} r(x,y';\phi).$
Therefore, the difference of reward functions for different answers $y,y'$ can be computed efficiently as the normalization constant $Z(x,\phi,\beta)$ does not appear, i.e.
\begin{align}
    r(x,y;\phi) -  r(x,y';\phi) = \beta \log \brr{\frac{\pi_\phi^\star(y|x)}{\pi_{\mathrm{ref}}(y|x)}} - \beta \log \brr{\frac{\pi_\phi^\star(y'|x)}{\pi_{\mathrm{ref}}(y'|x)}}. \label{eq:closed_form:reward_difference}
\end{align}
It follows that we can plug in the above analytical solution into the loss $\ell_{\mathrm{RM}}$ \eqref{eq:RLHF_loss} to obtain the following bi-level optimization problem that finds the optimal reward parameterization:
 \begin{subequations}
\label{formulation:rlhf_equivalence}
    \begin{align}
    \min_{\phi \in \Phi}\ \ell(\phi)
    &:=- \mathbb{E}_{x \sim \rho, (y_l\prec y_w)\sim\pi^{P}(\cdot|x)} \Big[ \log \Big( \sigma \big( \beta \log \brr{\frac{\pi^\star_\phi(y_w | x)}{\pi_{\mathrm{ref}}(y_w |x)}} - \beta \log \brr{\frac{\pi^\star_\phi(y_l | x)}{\pi_{\mathrm{ref}}(y_l | x)}} \big) \Big) \Big]   \label{eq:rlhf_equivalence:policy_objective} \\
    {\rm s.t.} \quad \pi^\star_\phi \  &:=\argmax_{\pi\in \Pi} ~~ \mathbb{E}_{x \sim \rho, y \sim \pi(\cdot|x)}\left[r(x, y;\phi)\right]  -{\beta}\mathbb{E}_{x \sim \rho}[D_{\mathrm{KL}}(\pi\left(\cdot | x\right)\|\pi_{\mathrm{ref}}\left(\cdot | x\right))], \label{eq:rlhf_equivalence:policy_constraint}
\end{align}
\end{subequations}
where $ \pi_\phi^\star $ is an optimal policy defined in \eqref{eq:rlhf_optimal_policy} under a certain reward model parameterized by $ \phi $. If one faithfully follows the above approach, then one would first compute the optimal reward $r(\cdot;\phi^*)$, then use it to identify the optimal policy using \eqref{eq:rlhf_optimal_policy}. To avoid training two separate models, \cite{rafailov2024direct} proposes to directly parameterize the policy and find the parameters $\theta^\star$ that solve the following problem: 
 \begin{align}\label{eq:DPO_loss}
    \argmin_{\theta \in \Theta}\ \ell_{\rm DPO}(\theta) :=- \mathbb{E}_{x \sim \rho, (y_l\prec y_w)\sim\pi^{P}(\cdot|x)} \bs{ \log \brr{ \sigma \brr{ \beta \log \brr{\frac{\pi(y_w|x;\theta)}{\pi_{\mathrm{ref}}(y_w|x)}} - \beta \log \brr{\frac{\pi(y_{l}|x;\theta)}{\pi_{\mathrm{ref}}(y_{l}|x)}}  } }}. 
\end{align}
Notice that, as shown in Protocol~\ref{prot:DPO}, an empirical estimate of this loss can be computed without the need of the dataset $\mathcal{D}_{\rm prompts}$ and online generations.
A first clear advantage of DPO over RLHF is that only the policy parameter is kept in memory. The reward network does not need to be stored, therefore, as shown in the Fig. \ref{fig:memory}, DPO is more memory efficient than RLHF. Further, no self-generation such as line 3 of Protocol \ref{prot:PPO} and \ref{prot:REINFORCE} is needed, so DPO is much easier to implement as compared to RL based algorithms.

\begin{wraptable}{l}{6cm}
    \centering \footnotesize
    \begin{tabular}{c|ccc}
    \toprule\small
         Action & $y_1$ & $y_2$ & $y_3$\\
         \midrule
         $\pi_{\mathrm{ref}}$ & $0.5$ & $0.5$ & $0$ \\
         $D_\mathrm{pref}$ & \multicolumn{3}{c}{$\{(y_w=y_1,y_l=y_2)\}$} \\
         \midrule
         $\pi_\mathrm{DPO}$ & $\alpha$ & $0.0$ & $1 - \alpha$ \\
         \midrule
         $\pi_\mathrm{RLHF}$ & 1 & 0 & 0 \\
         \bottomrule
    \end{tabular}
    \caption{\footnotesize The policy $\pi_{\mathrm{DPO}}$ minimizes $\ell_{\mathrm{DPO}}$ for any $\alpha \in [0,1)$. Hence, for $\alpha > 0$, it  produces a policy with support which is not a subset of $\pi_{\mathrm{ref}}$. The same can not happen for $\pi_{\mathrm{RLHF}}$ that minimizes $\ell_{\mathrm{RLHF}}$.}
    \vspace{0.2cm}
    \label{tab:dpo-counterexample}
\end{wraptable}

However, dropping the analytical form of $\pi_\phi^\star(y|x)$ in the last step of the DPO derivation creates practical differences between the performance of DPO and RLHF.
Therefore, DPO and RLHF are not equivalent, in particular, since we dropped a constraint, the solution set for DPO is a superset of the solution set for RLHF (Theorem 4.1 in \cite{xu2024dpo}). Recent studies suggest that RL-based fine-tuning has greater potential to enhance the performance of LLMs compared to the DPO \cite{xu2024dpo}
\begin{protocol}[t]
\caption{DPO (Direct Preference Optimization) \label{prot:DPO}}
    \centering
  \begin{algorithmic}[1]
  \State The learner receives as input: 
\begin{itemize}\item A preference dataset $\mathcal{D}_{\rm pref} = \{(x^i,y^i_w, y^i_\ell)\}^{N}_{i=1}$.
\item A policy function parameters class $\Theta$.
\end{itemize}
  \State The learner computes the stochastic loss.$$\widehat{\ell}_{\mathrm{DPO}}(\theta):= -
  \mathbb{E}_{(x ,y_w, y_l) \in \mathcal{D}_{\rm pref}} \bs{ \log \brr{ \sigma \brr{ \beta\log \brr{\frac{\pi(y_w|x;\theta)}{\pi_{\mathrm{ref}}(y_w|x)}} - \beta \log \brr{\frac{\pi^\star(y_{l}|x;\theta)}{\pi_{\mathrm{ref}}(y_{l}|x)}}  } }}$$
  \State The learner outputs $\pi_{\theta^\star}$ with $\theta^\star = \argmin_{\theta\in\Theta} \widehat{\ell}_{\mathrm{DPO}}(\theta)$.
  \end{algorithmic}
\end{protocol}

\begin{wraptable}{r}{7cm}
\centering
\footnotesize 
\begin{tabular}{c|ccc}
\toprule
                    & PPO Win & Tie & DPO Win \\ \midrule
PPO V.S. DPO   & 42      & 28  & 30       \\
\bottomrule
\end{tabular}
\caption{\footnotesize On HH-RLHF, GPT-4 is used to decide the winner between PPO and DPO models' output. Taken from \cite{xu2024dpo}.\label{tab:PPOvsRLHF}}
\end{wraptable}

From the formulation of DPO in Eq. \eqref{eq:DPO_loss}, we see that DPO only fits the given preference data and does not leverage online generations from its own model (see line 3 of Protocol \ref{prot:PPO} and \ref{prot:REINFORCE}). Due to the lack of the optimal policy constraint, the model trained by the DPO method can suffer from the limited generalization capability and can encounter the distribution shift issue when the coverage of the preference dataset is limited. 

Furthermore, when there exists distribution shift between the model outputs and the preference dataset, the performance of DPO can be significantly affected. 
A possible way to circumvent this problem is to use DPO in an online fashion \cite{qi2024online}. In contrast, in the RLHF training pipeline, the policy model is trained by RL and has leveraged online generations from the model itself, which can alleviate the distribution shift issue between model outputs and the preference dataset through leveraging the generalization power from the explicit estimated reward model. 
Similar empirical observations for the potential drawbacks of DPO compared with RLHF / PPO has also been observed in recent study \cite{xu2024dpo}. Hence, to unlock the full potential of LLMs, it appears that  RL is a necessary step in the post-training pipeline.  

The toy example in \cite{xu2024dpo}
that we report in Table~\ref{tab:dpo-counterexample} shows that the solution set of DPO is a strict superset of the solution set of RLHF. In particular, there are minimizers of the DPO loss  assign positive probability to answers for which no information is available ($y_3$ in this example).
This example shows that DPO should not be applied when the preference dataset has poor coverage of the answers set $\mathcal{Y}$.
For more practical problems, it is also observed that DPO  can be less performant in terms of winrate against the model fine tuned via RLHF; see Table~\ref{tab:PPOvsRLHF}.

\subsection{Learning from preference without the BT assumption.}
{ Regardless of using RLHF and the DPO approach to perform alignment, one key step is to use the BT model \eqref{eq:RLHF_loss} to model human preference. 
The BT assumption implies transitivity. This is restrictive because the preference dataset collected from different individuals might not be transitive even if each person follows a transitive model in generating the preference.
As an example, let us consider $3$ individuals $h_1, h_2, h_3$ and $3$ possible answers $y_1, y_2, y_3$. Let us denote $P_h$ the preference model of individuals $h$. Suppose that they follow the following preference models:
\begin{align*}
&\mathcal{P}_{h_1}(y_1 \succ y_2) = 1 \quad \mathcal{P}_{h_1}(y_2 \succ y_3) = 0 \quad \mathcal{P}_{h_1}(y_3 \succ y_1) = 1 \\
&\mathcal{P}_{h_2}(y_1 \succ y_2) = 0 \quad \mathcal{P}_{h_2}(y_2 \succ y_3) = 1 \quad \mathcal{P}_{h_2}(y_3 \succ y_1) = 1 \\
&\mathcal{P}_{h_3}(y_1 \succ y_2) = 1 \quad \mathcal{P}_{h_3}(y_2 \succ y_3) = 1 \quad \mathcal{P}_{h_3}(y_3 \succ y_1) = 0.
\end{align*}
Notice, that each of these models is transitive. However, the average model defined as $\mathcal{P}(y \succ y') = \frac{1}{3} \sum_{h \in \{ h_1, h_2, h_3\}} \mathcal{P}_h(y \succ y')$ satisfies $\mathcal{P}(y_1 \succ y_2) = \mathcal{P}(y_2 \succ y_3) = \mathcal{P}(y_3 \succ y_1) = 2/3$. That is, the average model is non transitive and can not be modeled by the BT assumption.
In the sense, the BT assumption is data wasteful. In this example, one should consider preferences only from a single individual in order to make the BT assumption valid.
Not enforcing the BT assumption, allows to use more data, i.e. preferences from all the three individuals.
Given this motivation, recent works \cite{munos2023nash} investigated learning from preferences directly. 
 The training happens in two stages as in RLHF: the first step uses maximum likelihood to learn a preference model rather than a reward function, and a second step identifies  a Nash equilibrium rather than an optimal policy. 
\paragraph{Preference learning}
First, we learn via maximum likelihood, 
a preference model $\mathcal{P}: \mathcal{X} \times \mathcal{Y} \times \mathcal{Y} \rightarrow [0,1]$ such that $\mathcal{P}(y \succ y' | x)$ is the probability that certain oracle, which produced the preferences in the dataset $\mathcal{D}_{\rm pref}$, prefers answer $y$ over $y'$ following the prompt $x$.
\begin{protocol}[t]
\caption{NLHF (Nash Learning from Human Feedback) \label{prot:NLHF}}
    \centering
  \begin{algorithmic}[1]
  \State The learner receives as input: 
(i) A preference dataset $\mathcal{D}_{\rm pref} = \{(x^i,y^i_w, y^i_\ell)\}^{N}_{i=1}$, (ii) a prompt dataset $\mathcal{D}_{\mathrm{prompts}} = \{ x^i\}^{N_{\mathrm{prompts}}}_{i=1}$, (iii) a preference model function parameters class $\Phi$ and (iv) a policy function parameters class $\Theta$.
\State The learner estimates the preference model as $\mathcal{P}\brr{\cdot \succ\cdot|\cdot;\phi^\star}$ with 
\begin{equation*}
    \phi^\star = \argmin_{\phi\in\Phi}\ \widehat{\ell}_{\rm PM}(\phi)
    := - \mathbb{E}_{(x,y_w,y_l)\in \mathcal{D}_{\rm pref}} \log \Big( \mathcal{P}( y_w \succ y_l | x; \phi) \Big).
\end{equation*}
  \State The learner computes the stochastic objective
  \begin{align*}
  \widehat{\ell}_{\mathrm{NLHF}}(\theta, \theta') :=& -\mathbb{E}_{x\in\mathcal{D}_{\rm prompts}, y \sim \pi(\cdot|x;\theta), y' \sim \pi(\cdot|x;\theta')}\mathcal{P}(y \succ y'| x; \phi^\star) \\ &+ \beta D_{\mathrm{KL}}(\pi\left(\cdot | x;\theta\right)\|\pi_{\mathrm{ref}}\left(\cdot | x\right)) - 
  \beta D_{\mathrm{KL}}(\pi\left(\cdot | x;\theta' \right)\|\pi_{\mathrm{ref}}\left(\cdot | x\right)),
  \end{align*}
  \State The learner outputs $\pi_{\theta^\star}$ with $\theta^\star = \argmin_{\theta\in\Theta} \max_{\theta' \in \Theta}\widehat{\ell}_{\mathrm{NLHF}}(\theta, \theta')$.
  \end{algorithmic}
\end{protocol}
Mirroring the procedure in RLHF, let us consider a preference model class with parameters class $\Phi$, i.e. $\bcc{\mathcal{P}(\cdot \succ \cdot | \cdot; \phi)}_{\phi \in \Phi}$, consider the following minimization problem (where ${\rm PM}$ stands for preference modelling): 
\newline \noindent 
$
\phi^\star = \argmin_{\phi \in \Phi}\ \ell_{\rm PM}(\phi)
    :=- \mathbb{E}_{x \sim \rho, (y_l\prec y_w)\sim\pi^{P}(\cdot|x)} \Big[ \log \Big( \mathcal{P}(y_w \succ y_l | x; \phi) \Big) \Big]. $
Protocol~\ref{prot:NLHF} shows how to do this practically, minimizing an empirical unbiased estimate for $\ell_{\rm PM}$. 
\paragraph{Nash equilibrium learning}
After learning the preference model, \cite{munos2023nash} proposes to find the unique  Nash equilibrium of the objective $\widehat{\ell}_{\rm NLHF}$ in Protocol~\ref{prot:NLHF}. Due to the min max structure of this problem, the optimization algorithm design is more delicate and require careful conservative updates to ensure last iterate convergence.
However, this additional complexity on the optimization side comes at the benefit of dropping the BT assumption and therefore accommodating a larger number of preference data.

Experiments in \cite[Table 1]{munos2023nash} shows that the NLHF method dubbed MD1 achieves win rate of $0.598$ against the same model aligned via RLHF on a text summarization task.
The drawback of this method is that it requires solving a minmax optimization problem rather than the simpler minimization problem required in RLHF.} 

\subsection{Learning from data  verifiers via Rule-Based Reinforcement Learning}\label{sub:verifier}

Before closing this section, we would like to mention a recently developed alignment strategy, which is fundamental for training powerful reasoning models such as DeepSeek-R1 \cite{guo2025deepseek}. The general idea is simple: To teach the model to solve many challenging math, STEM or coding  related problems, a reward model is no longer needed. Rather, one can simply use 
a \emph{solution verifier}, which simply verifies whether the solution is correct or not.
In \cite{guo2025deepseek}, the authors proposed to use LeetCode tests case  as verifier for programming questions, while creating a dataset of mathematical problems from OpenWebMath 
for math verifiers.  
In particular, in the reproduction of the Deepseek R1, \cite{xie2025logic} adopted the rule-based reward system that comprises two types of rewards: Format Reward and Answer Reward. Here, we show the detailed reward designed in \cite{xie2025logic} as below:
{\small
\begin{align}
r_{\rm format} =
\begin{cases} 
1, & \text{if format is correct} \\
-1, & \text{if format is incorrect}
\end{cases}, \quad
r_{\rm answer} =
\begin{cases} 
2, & \text{if the answer fully matches the ground truth} \\
-1.5, & \text{if the answer partially mismatches the ground truth} \\
-2, & \text{if the answer cannot be parsed or is missing}
\end{cases} \nonumber
\end{align}
}
where $r_{\rm format}$ is the format reward and $r_{\rm answer}$ is the answer reward.

However in \cite{guo2025deepseek}, it is noted that the model which optimizes purely the rule based reward, i.e. DeepSeek-R1-Zero leads to poor language skills suffering for example from language mixing. To avoid this, \cite{guo2025deepseek} uses multi-stage training with learning with verifiable rewards at the initial stage and SFT/preference based learning used at the final stages to improve the language skills of the model.
In the initial stage, the rule based reward is optimized via GRPO in Algorithm~\ref{prot:GRPO}.
\section{A Unified Perspective for LLM Alignment}\label{sec:unified}
So far we have discussed a number of state-of-the-art approaches for LLM alignment. Generally speaking, they can be divided into two categories: alignment using the demonstration data, and alignment using the preference data. One might wonder why algorithms belonging to these two categories are so different (e.g., supervised learning vs. reward learning + RL). After all, both families of algorithms are trying to learn from some forms of human preference. 
In this section, we provide a perspective derived from the stochastic choice model discussed in Section \ref{sec:theoretical}, which serves to unify different approaches of alignment. 


Recall that in Section \ref{sec:theoretical}, we have introduced two estimation problems \eqref{eq:likelihood:rm} and \eqref{eq:likelihood:sft}. Although they use different datasets (i.e., the former uses the preference dataset $\mathcal{D}{\rm pref}$, while the latter uses the demonstration dataset $\mathcal{D}{\rm demo}$), they share a few common properties: (1) the reward function can be learned from both datasets and (2) both maximize some form of the likelihood function. These observations lead to the following questions:
\begin{itemize}
\item (Q1) Given that the formulation of the demonstration-based problem also allows for reward learning, why do existing alignment algorithms only learn rewards from the preference dataset but not from the demonstration dataset?
\item (Q2) Since both problems maximize the likelihood functions, why do we need to separate the alignment process into different stages (i.e., SFT and RLHF)?
\end{itemize}
In the following, we will first address question (Q1) by formulating a new problem that learns {\it both} the reward and the policy from $\mathcal{D}_{\rm demon}$. We will show that this formulation is closely related to a class of RL problem called inverse RL (IRL), and we will derive efficient methods from the IRL literature. Somewhat unexpectedly, it turns out that a number of algorithms discussed in the previous section, such as SFT and SPIN,  can be viewed as special cases of the IRL-based approach. 

Furthermore, leveraging the development in (Q1), we address (Q2) by introducing a framework that unifies the two stages of alignment problem into a {\it single} stage. The new framework,  named Alignment with Integrated Human Feedback (AIHF), jointly learns a reward and a policy by maximizing the likelihood of {\it both} $\mathcal{D}_{\rm demon}$ and $\mathcal{D}_{\rm pref}$. We will discuss the connection of AIHF with various alignment algorithms introduced in the previous section.



\subsection{Jointly learning policy and reward from the demonstrations data}


To jointly learn a reward and a policy from the demonstration dataset $\mathcal{D}_{\rm demon}$, let us leverage the estimation problem \eqref{eq:likelihood:sft} derived from the stochastic choice theory. Given the demonstration $(x, y)$ sampled from an expert policy $\pi^{\rm E}$, we can directly translate \eqref{eq:likelihood:sft} with the optimal reward defined in \eqref{eq:pi:star}, into the LLM alignment setting and obtain the following problem \cite{zeng2022maximum}:
\begin{subequations} \label{formulation:IRL}
    \begin{align}
    \max_{\phi} \ &\ell_{\mathrm{IRL}}(\phi):= \mathbb{E}_{x \sim \rho, y \sim \pi^{\rm E}(\cdot | x)}\left[\log \pi^\star_{r_{\phi}}\left(y | x\right)\right] \label{eq:obj:irl} \\
    \pi^\star_{r_{\phi}} &:=  \argmax_{\pi} ~~ \mathbb{E}_{x \sim \rho, y \sim \pi(\cdot|x)}\left[r(x, y;\phi)\right]  -{\beta}\mathbb{E}_{x \sim \rho}[D_{\mathrm{KL}}(\pi\left(\cdot | x\right)\|\pi_{\mathrm{ref}}\left(\cdot | x\right))]. \label{eq:constraint:irl}
\end{align}
\end{subequations}
 Compared with the SFT formulation in \eqref{eq:SFT}, we see that \eqref{formulation:IRL} shares the same objective, but has an additional constraint which ensures that the policy has to be an optimal policy under a certain reward model.

The problem \eqref{formulation:IRL} falls under the class of inverse reinforcement learning (IRL) problems, where the objective is to jointly learn an optimal policy and the underlying reward model, assuming access to expert policy samples. Similarly, as noted in Sec. \ref{sec:cloning}, the conventional SFT formulation \eqref{eq:SFT} is a form of BC that aims to replicate an expert’s policy. However, BC generally exhibits poorer generalization than IRL. Indeed, \cite{ross2011reduction} shows that BC is a suboptimal approach to learning from demonstrations, leading to regret bounds that scale quadratically with the problem horizon. This arises due to distribution shift -- the learned policy may encounter states not covered in the demonstration data, causing compounding errors that increasingly deviate from the expert trajectory over time.

In contrast, IRL mitigates distribution shift and achieves better generalization by explicitly modeling the expert’s reward function, rather than simply mimicking actions. By recovering an estimate of the reward structure that drives expert behavior, IRL enables the agent to make goal-directed decisions even in unseen states. This contrasts with BC, which lacks an underlying objective thus often struggles in novel situations. Additionally, optimizing policies with a learned reward function allows for adaptive behavior, where the agent can deviate from expert demonstrations in a way that still aligns with the inferred intent. As a result, IRL-based policies are generally more robust in complex environments where state distributions differ from training data \cite{viano2021robust}. {Recognizing the limitations of standard SFT methods, particularly regarding distribution shift, researchers have proposed a range of IRL-based techniques \cite{wu2021textgail,li2024getting,wulfmeier2024imitating,zeng2025demonstrations} to more effectively leverage demonstration data.}

Unfortunately, the IRL problem \eqref{formulation:IRL} appears to be much more involved compared to the SFT problem \eqref{eq:SFT}, as it involves {\it two} levels of optimization problem stacked in hierarchy. It turns out that there is a way to simplify it. In \cite{zeng2024structural, zeng2025from}, the authors show that when a finite demonstration dataset $\mathcal{D}_{\text{demo}} := \{ (x, y) \}$ is available,  one can consider the following surrogate objective: 
\begin{align}
	\max_{\phi} \min_{\theta} ~ \widetilde{L}(\phi, \theta; \mathcal{D}_{\rm demo}) :=
 \mathbb{E}_{(x, y) \sim \mathcal{D}_{\rm demo}, y' \sim \pi(\cdot | x; \theta )} \Big[ r(x, y; \phi) -  r(x, y'; \phi) + \beta {D}_{\rm KL}\Big( \pi(\cdot | x; \theta) \| \pi_{\text{ref}}(\cdot | x) \Big) \Big]. \label{ML:estimation:surrogate} 
\end{align}
{As a remark, one can plug the closed-form expression of the optimal policy $ 
\pi^*_{r_{\phi}} $ into the maximum likelihood objective \eqref{eq:obj:irl}, then one can see that the surrogate objective \eqref{ML:estimation:surrogate} is a finite-sample version of the IRL problem \eqref{formulation:IRL}. }
Further, as the number of demonstration data becomes large, problem \eqref{ML:estimation:surrogate} becomes asymptotically close to problem \eqref{formulation:IRL}, in the sense that their optimal solutions become asymptotically close to each other (with provable bounds) 
\cite{zeng2024structural}. {The surrogate objective in \eqref{ML:estimation:surrogate} offers an alternative interpretation of the IRL problem formulated in \eqref{formulation:IRL} from the adversarial training perspective. Specifically, the reward estimator is optimized to distinguish between expert demonstrations and self-generated trajectories (the max problem), while the policy is optimized to generate trajectories which can best mimic the expert demonstrations and the self-generated trajectories (the min problem).} Therefore, it is reasonable to develop practical algorithms based on \eqref{ML:estimation:surrogate}, which is apparently much easier to optimize.  
As suggested in \cite{zeng2025from}, one can 
use the following 
alternating algorithm to optimize the reward and the policy:
\begin{subequations} \label{irl:algorithmic_pipeline}
    \begin{align}
    \phi_{t+1} &:= \argmin_{\phi}  ~ - \mathbb{E}_{(x, y) \sim \mathcal{D}_{\rm demo}, y' \sim \pi(\cdot | x; \theta_{t} )} \Big[ \ell \Big( r(x, y; \phi) -  r(x, y'; \phi) \Big) \Big]. \label{update:pipeline:reward} \\
        \theta_{t+1} &:= \argmax_{\theta} ~~ \mathbb{E}_{x \sim \rho, y \sim \pi(\cdot|x;\theta)}\left[r(x, y;\phi_{t+1})\right]  -{\beta}\mathbb{E}_{x \sim \rho}[D_{\mathrm{KL}}(\pi\left(\cdot | x; \theta \right)\|\pi_{\mathrm{ref}}\left(\cdot | x\right))]. \label{update:pipeline:policy} 
    \end{align}
\end{subequations}
Note that if we directly apply the above gradient descent/ascent algorithm to \eqref{ML:estimation:surrogate}, then $\ell(\cdot)$ in \eqref{update:pipeline:reward} should be a linear function $\ell(x) = x$. Considering that the use of a linear loss function results in an unbounded objective value, one can typically choose some easy-to-optimize loss functions such as  log sigmoid function $\ell(z) := \log \sigma (z)$. 
Next, let us make some remarks about the algorithm \eqref{irl:algorithmic_pipeline}.

\emph{Connection to the RLHF.} We observe that  \eqref{irl:algorithmic_pipeline} is an iterative update scheme which {\it integrates} the reward and policy updates. 
This closely resembles the RLHF pipeline, where both the reward and policy are updated. There are two major differences. First, in RLHF the reward and the policy are updated in two {\it separate} stages, while here they are updated {\it iteratively}. Second, since only demonstration data $\mathcal{D}_{\rm demon}$ are available, when learning a reward model in \eqref{update:pipeline:reward}, the demonstration is always treated as the preferred data, while the non-preferred data are  the model-generated responses. This is reasonable, under the assumption that expert data are of high quality. {In practice, RLHF-based and IRL-based methods are complementary to each other. It’s common to leverage both demonstration and preference datasets jointly in the training pipeline to achieve better performance \cite{ibarz2018reward,tunstall2023zephyr}. In \cite{li2024joint, xiao2025connection}, the connections has been shown that both IRL and RLHF are imitating a target policy using different dataset (demonstrations and preferences). Leveraging this insight, the alignment pipeline can incorporate both demonstration and preference datasets, improving model performance by integrating RLHF with IRL techniques.  In general, it can be shown that RLHF could be a special form of IRL. Here, we show that IRL models the preference of a single action over all possible actions in \eqref{eq:obj:irl} when maximizing the likelihood of one demonstration action, whereas RLHF models the relative preference between pairs of actions.

Below we show that the reward modeling in RLHF can be derived from the optimal policy expression in \eqref{eq:rlhf_optimal_policy}.
As we have defined in \eqref{eq:rlhf_optimal_policy}, given the reward parameter $\phi$, the optimal policy of the regularized RL problem defined in \eqref{eq:RLFT} has the closed-form $ \pi^\star_{r_{\phi}}(y | x) = \frac{\pi_{\mathrm{ref}}(y|x)\exp(\beta^{-1} r(x,y;\phi))}{Z(x,\phi,\beta)} $. With this expression, we can now obtain a model for the likelihood that sequence $y_w$ is preferred over $y_l$. 
With a pair of actions $\{y_l, y_w\}$, by the {\em independence of irrelevant alternatives} property \cite{fudenberg2015stochastic} of the optimal choice $\pi^*_{r_{\phi}}$, the likelihood that sequence $y_w$ is preferred over $y_l$ is given by $\mathbb{P}_{\pi^*_{r_{\phi}}}(y_w  \succ y_l | x):=\frac{\pi^*_{r_{\phi}}(y_w | x)}{\pi^*_{r_{\phi}}(y_w | x) + \pi^*_{r_{\phi}}(y_l | x)}.$ 
This motivates the choice model as the following {\it likelihood function}:
{\small \begin{align*}
L(R(\cdot;\phi))  &= \mathbb{E}_{(y_w  \succ y_l)} \Big[ \log \frac{\pi^*_{r_{\phi}}(y_w | x)}{\pi^*_{r_{\phi}}(y_w | x) + \pi^*_{r_{\phi}}(y_l | x)} \Big]\nonumber \\ 
& = \mathbb{E}_{(y_w  \succ y_l)} \Big[ \log\frac{\pi_{\mathrm{ref}}(y_w|x)\exp(\beta^{-1} r(x,y_w;\phi))}{\pi_{\mathrm{ref}}(y_w|x)\exp(\beta^{-1} r(x,y_w;\phi)) + \pi_{\mathrm{ref}}(y_l|x)\exp(\beta^{-1} r(x,y_l;\phi))} \Big].
\end{align*}}%
When $\pi_{\mathrm{ref}}$ equals to the uniform distribution, this model is equivalent to
the BTL model in RLHF:
\begin{align}
L(R(\cdot;\phi)) = \ell_{\rm RM}(\phi)
    :=- \mathbb{E}_{(y_w\succ y_l)} \Big[ \log \Big( \sigma \big( r(x, y_w; \phi) - r(x, y_l;\phi) \big) \Big) \Big].  \label{preference_loss:BTL}
\end{align}
For the IRL formulation defined in \eqref{formulation:IRL}, we see that it models the  preference of a single action over the entire action space. When the action space consists of only two actions, we see that RLHF is a special case of IRL and the reward modeling between IRL and RLHF is equivalent.
}


\emph{Connection to SPIN.} There is also an interesting connection between \eqref{ML:estimation:surrogate} and the SPIN introduced in Sec. \ref{sub:spin}. Note that in \eqref{eq:closed_form:reward_difference}, we have shown that the reward difference can be expressed by using the definition of the optimal policy. By plugging \eqref{eq:closed_form:reward_difference} into \eqref{update:pipeline:reward}, we obtain the following problem:
\begin{align}
    \min_{\phi \in \Phi}\ \ell(\phi)
    &:=- \mathbb{E}_{(x, y) \sim \mathcal{D}, y' \sim \pi(\cdot | x; \theta_{t} )} \Big[ \ell \Big( \beta \log \brr{\frac{\pi^\star_{r_{\phi}}(y | x)}{\pi_{\mathrm{ref}}(y |x)}} - \beta \log \brr{\frac{\pi^\star_{r_{\phi}}(y' | x)}{\pi_{\mathrm{ref}}(y' | x)}} \Big) \Big]   \nonumber \\
    s.t. \quad \pi^\star_{r_{\phi}} \  &:=\argmax_{\pi} ~~ \mathbb{E}_{x \sim \rho, y \sim \pi(\cdot|x)}\left[r(x, y;\phi)\right]  -{\beta}\mathbb{E}_{x \sim \rho}[D_{\mathrm{KL}}(\pi\left(\cdot | x\right)\|\pi_{\mathrm{ref}}\left(\cdot | x\right))]. \nonumber
\end{align}
Similarly, as discussed in the DPO approach, one can make two simplifications (1) remove the constraints that the policy has to be optimal w.r.t. some reward; (2) directly parameterize the resulting policy using $\theta$. Then the above problem will be reduced precisely to the SPIN formulation \eqref{eq:spin}. 

To further compare the SPIN and IRL, note that at each iteration, once the synthetic preference dataset has been constructed where the preferred data are the demonstration / SFT data and the nonpreferred data are the model-generated data, SPIN directly leverages the synthetic preference dataset and runs DPO to update the policy,  while the IRL approach considers \textit{two} separate steps to update \textit{both} the reward model and the policy model. Similarly to the comparison of DPO and RLHF, here we expect that IRL enjoys better generalization and can alleviate the distribution shift issue through leveraging online generations in the RL training step, at the cost of a heavier memory burden and additional online generations.

In summary, the method discussed in this section addressed the question (Q1) we raised at the beginning of this section --  instead of directly performing SFT on the demonstration data $\mathcal{D}_{\rm demon}$, it is also possible to learn a reward function from such a dataset, and we expect that the resulting model will generalize better. 
Later we will use numerical results to confirm such an advantage. 

\subsection{Jointly learning policy and reward from Integrated Human Feedback}


Next, let us address question (Q2), by developing an approach that {\it integrates} different steps of the 
alignment process. Toward this end, we observe that by leveraging the IRL-based formulation \eqref{formulation:IRL}, learning from the demonstration data $\mathcal{D}_{\rm demon}$ becomes very similar to the standard RLHF approach discussed in Sec. \ref{sub:preference:prompts}, which learns preference data $\mathcal{D}_{\rm pref}$ -- both maximize the likelihood of observing the data, while explicitly learning a reward function. Therefore, using reward learning as a common component, it is straightforward to integrate these problems together. In \cite{li2024joint} the authors considered the following problem, which extends the IRL problem  \eqref{formulation:IRL} by adding the BT loss \eqref{eq:RLHF_loss}:
\begin{subequations} \label{formulation:AIHF}
    \begin{align}
    \max_{\phi} \ &\ell_{\mathrm{AIHF}}(\phi):= \gamma \mathbb{E}_{x \sim \rho, y \sim \pi^{\rm E}(\cdot | x)}\left[\log \pi^\star_{r_{\phi}}\left(y | x\right)\right] \nonumber\\
    & \quad \quad \quad \quad \quad + \alpha \mathbb{E}_{x \sim \rho, (y_l\prec y_w)\sim\pi^{P}(\cdot|x)} \Big[ \ell \Big( r(x, y_w; \phi) - r(x, y_l;\phi)  \Big) \Big] \label{aihf:formulation:reward}\\
    \pi^\star_{r_{\phi}} &:=  \argmax_{\pi} ~~ \mathbb{E}_{x \sim \rho, y \sim \pi(\cdot|x)}\left[r(x, y;\phi)\right]  -{\beta}\mathbb{E}_{x \sim \rho}[D_{\mathrm{KL}}(\pi\left(\cdot | x\right)\|\pi_{\mathrm{ref}}\left(\cdot | x\right))], \label{aihf:formulation:policy}
\end{align}
\end{subequations}
where $\gamma\ge 0$ and $\alpha\ge 0$ are two tuneable parameters. 
The authors refer to this approach as AIHF (Alignment with Integrated Human Feedback).

To design an algorithm to solve problem \eqref{formulation:AIHF}, one can combine the RLHF Protocol \ref{prot:RLHF} and the IRL algorithm in \eqref{irl:algorithmic_pipeline}, to obtain the following iterative algorithm (where 
$t$ denotes the iteration number): 
\begin{subequations} \label{AIHF:Algorithmic_Pipeline}
    \begin{align}
    \phi_{t+1} &:= \argmin_{\phi}  ~ - \gamma \mathbb{E}_{(x, y) \sim \mathcal{D}_{\rm demo}, y' \sim \pi(\cdot | x; \theta_{t} )} \Big[ \ell \Big( r(x, y; \phi) -  r(x, y'; \phi) \Big) \Big] \nonumber \\
    & \quad \quad \quad \quad \quad \quad \quad \, \, \, - \alpha \mathbb{E}_{(x, y_w, y_l) \sim \mathcal{D}_{\rm pref}} \Big[ \ell \big( r(x, y_w; \phi) - r(x, y_l;\phi) \big) \Big], \label{aihf:pipeline:reward} \\
        \theta_{t+1} &:= \argmax_{\theta} ~~ \mathbb{E}_{x \sim \rho, y \sim \pi(\cdot|x;\theta)}\left[r(x, y;\phi_{t+1})\right] - {\beta}\mathbb{E}_{x \sim \rho}[D_{\mathrm{KL}}(\pi\left(\cdot | x; \theta \right)\|\pi_{\mathrm{ref}}\left(\cdot | x\right))]. \label{aihf:pipeline:policy} 
    \end{align}
\end{subequations}
In fact, compared to the IRL algorithm 
\eqref{irl:algorithmic_pipeline}, the only difference is that the reward is learned from {\it both} preference dataset and the demonstration dataset. In \cite{li2024joint}, the authors observed that learning reward function from a diverse set of data can be more robust to situations, e.g. when there is an imbalance between the demonstration and preference data, resulting in a more robust policy model; see \cite[Sec.3.4]{li2024joint} for detailed discussion and a few examples showcasing such an advantage. 

Finally, we note that AIHF has several notable special cases. First, if one chooses $\gamma =0$, then reward learning and policy learning become {\it completely decoupled}, reducing to the orignal two-step RLHF approach discussed in Section \ref{sub:preference:prompts}.
Second, similarly to DPO, one can plug in the optimal solution of the lower-level problem (as given in \eqref{eq:closed_form:reward_difference}) to the upper-level, remove the lower-level constraint \eqref{aihf:formulation:policy}, and directly parameterize the policy. The resulting problem becomes:
\begin{equation*}
    \max_{\theta} \ \mathbb{E}_{x \sim \rho, y \sim \pi^{\rm E}(\cdot | x;\theta)}\left[\log \pi\left(y | x\right)\right] + \alpha \mathbb{E}_{x \sim \rho, (y_l\prec y_w)\sim\pi^{P}(\cdot|x)} \Big[ \ell \Big( \beta \log \brr{\frac{\pi(y_w | x; \theta)}{\pi_{\mathrm{ref}}(y_w |x)}} - \beta \log \brr{\frac{\pi(y_l | x; \theta)}{\pi_{\mathrm{ref}}(y_l | x)}} \Big) \Big].
\end{equation*}
This corresponds to the Regularized Preference Optimization (RPO) approach proposed in \cite{liu2024provably}.
Moreover, if we drop \eqref{aihf:pipeline:policy} and apply the DPO trick to \eqref{AIHF:Algorithmic_Pipeline} to optimize:
\begin{equation*}
\begin{aligned}
    \max_{\theta}  \ & \mathbb{E}_{(x, y) \sim \mathcal{D}_{\rm demo}, y' \sim \pi(\cdot | x; \theta_{t} )} \Big[ \ell \Big( \beta \log \brr{\frac{\pi(y | x; \theta)}{\pi_{\mathrm{ref}}(y |x)}} - \beta \log \brr{\frac{\pi(y' | x; \theta)}{\pi_{\mathrm{ref}}(y' | x)}} \Big) \Big] \\ &\quad \,\,  + \alpha \mathbb{E}_{(x, y_w, y_l) \sim \mathcal{D}_{\rm pref}} \Big[ \ell \Big( \beta \log \brr{\frac{\pi(y_w | x; \theta)}{\pi_{\mathrm{ref}}(y_w |x)}} - \beta \log \brr{\frac{\pi(y_l | x; \theta)}{\pi_{\mathrm{ref}}(y_l | x)}} \Big) \Big].
\end{aligned}
\end{equation*}
Then it corresponds to the  Self-Play with Adversarial Critic (SPAC) approach proposed in \cite{ji2024self}.

Finally, we conclude this section with a tabular comparison of different types of alignment algorithms.

\begin{table}[t]
\centering
\small
\begin{tabular}{@{}lccc@{}}
\toprule
\textbf{Algorithm} & \textbf{Needs Reward?} & \textbf{Dataset Type} & \textbf{Needs Value Fn.?} \\
\midrule
SFT & \xmark & Demonstration & \xmark \\
SPIN & \xmark & Demonstration + self-generated & \xmark \\
IRL & \cmark & Demonstration & \xmark \\
REINFORCE & \cmark & Demonstration + Preference & \xmark \\
RLOO & \cmark & Demonstration + Preference & \xmark \\
PPO & \cmark & Demonstration + Preference & \cmark \\
DPO & \xmark & Demonstration + Preference & \xmark \\
NLHF & \xmark~(pref.~model) & Preference & \xmark \\
GRPO & \cmark & Demonstration + Preference & \xmark \\
Rule-based RL & \cmark~(rule/verifier) & Verifiable tasks & \xmark \\
AIHF & \cmark & Demonstration + Preference & \textit{flexible} \\
\bottomrule
\end{tabular}
\caption{Comparison of alignment algorithms by reward requirement, dataset type, and value-function (critic) dependence. Here since AIHF can use different kinds of RL algorithms such as PPO and DPO, we denote it as `flexible' in the last column.}
\label{tab:alignment_summary}
\end{table}

Table~\ref{tab:alignment_summary} summarizes different alignment algorithms according to their data sources and optimization requirements. 
Demonstration-based methods (e.g., SFT, SPIN, IRL) rely solely on expert trajectories, whereas preference-based methods (e.g., DPO, PPO, RLOO) depend on pairwise comparisons or on demonstration data when initialized from an SFT model. 
Among these, PPO-style variants explicitly require a value function to compute advantages, while most direct preference-modeling approaches remove the need for critic estimation. 
Hybrid frameworks such as AIHF integrate both demonstration and preference data, effectively bridging supervised and reinforcement-based alignment paradigms.


\section{Dataset, Benchmarks, and Evaluation}\label{sec:evals}
In this section, we will introduce a few  datasets and benchmarks that are often used to evaluate alignment algorithms. Further, we will present some evaluation results comparing a number of algorithms we have discussed so far. 

\vspace{-0.5cm}
\subsection{Dataset}
Typically, datasets used in LLM alignment can be categorized into two types: (1) datasets containing multiple responses for a given prompt, with human (or AI-generated) rankings indicating preference, and (2) datasets containing a single demonstration response per prompt.

\noindent{\bf Preference Dataet}. Some of the most well-known preference datasets include: (1) HH-RLHF \footnote{Available at \url{https://huggingface.co/datasets/Anthropic/hh-rlhf}.}(Anthropic's Helpful-Harmless RLHF), a dataset collected by Anthropic that includes human preference annotations over pairs of responses, focusing on helpfulness (how well the response answers the query) and harmlessness (avoiding toxic, biased, or dangerous content). (2) UltraFeedback \footnote{Available at \url{https://huggingface.co/datasets/openbmb/UltraFeedback}.}, which consists of 64K prompts, each accompanied by four model completions from a diverse set of open and proprietary models. Preference labels are provided by GPT-4 based on criteria such as helpfulness and honesty. This dataset has been used to train state-of-the-art chat models such as Zephyr-7B-$\beta$. 

\noindent{\bf Demonstration Dataset}. Some of the most well-known demonstration datasets include: (1) the TL;DR dataset \footnote{Available at \url{https://huggingface.co/datasets/CarperAI/openai_summarize_tldr}.}, a dataset collected from Reddit where users provide short summaries (TL;DRs) of their posts. (2) UltraChat \footnote{Available at \url{https://huggingface.co/datasets/stingning/ultrachat}.}, an open-source, large-scale, multi-turn dialogue dataset generated using Turbo APIs. It contains 774K training samples and is designed to enhance instruction-following capabilities.

{
\noindent{\bf Self-Generated Dataset} In addition to using public datasets, another line of research leverages self-generated data for policy optimization. For example, given a prompt, \cite{abdolmaleki2024preference} samples one or more responses from a policy $\pi$, evaluates each input-generation pair using a evaluation function (e.g., a reward model in RLHF), labels them as preferred or dispreferred, and then updates the policy using the EM algorithm based on these individual preferences. Similarly, \cite{xiong2023iterative} also employs self-generated data to iteratively improve the policy model. This approach can lead to more uniform coverage of the prompt-response space. }

\vspace{-0.2cm}

\subsection{Benchmark datasets and benchmark models}
\noindent{\bf AlpacaEval}. AlpacaEval \cite{dubois2024length} is an LLM-based automated evaluation metric – it operates on a fixed set of 805 instructions chosen to be representative of user interactions on the Alpaca web demo. A GPT-4 turbo-based evaluator then compares the responses head-to-head and outputs the probability of preferring the evaluated model. The win rate is then computed as the expected probability that the auto-evaluator prefers the evaluated model’s output on the 805 instructions. This win rate serves as a performance
measure of the evaluated LM chatbot.

\noindent{\bf Open LLM Leaderboard}. The HuggingFace Open LLM Leaderboard framework \cite{open-llm-leaderboard} is widely adopted for evaluating LLMs. This comprehensive evaluation suite assesses LLM performance across six key tasks: commonsense reasoning, including ARC
, HellaSwag
, and Winogrande; 
multi-task language understanding, as measured by MMLU; truthfulness assessment, which evaluates a model’s tendency to mimic human falsehoods using TruthfulQA; and mathematical problem-solving, assessed via GSM8K.

\noindent{\bf MT-Bench}. MT-Bench \cite{zheng2023judging} is a benchmark consisting of 80 high-quality multi-turn questions, specifically designed to evaluate multi-turn conversation and instruction-following capabilities in LLMs. It focuses on common real-world use cases while incorporating challenging questions to effectively differentiate model performance. To ensure comprehensive coverage, the benchmark is structured around eight key categories of user prompts: writing, roleplay, extraction, reasoning, mathematics, coding, knowledge I (STEM), and knowledge II (humanities/social sciences).

{\noindent{\bf HELM} Holistic Evaluation of Language Models (HELM) \cite{liang2023holistic} consists of two levels: (i) an abstract taxonomy of scenarios and metrics that defines the design space for language model evaluation, and (ii) a concrete set of implemented scenarios and metrics selected to prioritize coverage (e.g., different English varieties), value (e.g., user-facing applications), and feasibility (e.g., limited engineering resources). HELM currently includes a core set of 16 scenarios and 7 metrics. These scenarios cover six user-facing tasks (e.g., question answering, information retrieval, summarization, toxicity detection) and span multiple domains (e.g., news, books).

\noindent{\bf Artificial Analysis\footnote{Available at \url{https://artificialanalysis.ai/leaderboards/models}.}} Artificial Analysis provides objective evaluations and insights to assist developers, researchers, businesses, and other AI users in selecting the most suitable AI technologies for their specific use cases. This benchmark compares and ranks the performance of over 30 LLMs across key metrics, including output quality, cost, performance, and speed, as well as context window size and other relevant factors.
}

\noindent{\bf Reasoning Benchmark}. Several challenging benchmarks are also used to evaluate reasoning ability. For example, mathematical benchmarks such as AIME (American Invitational Mathematics Examination), AMC (American Mathematics Competitions), and OlympiadBench \cite{he2024olympiadbench} assess problem-solving skills in competitive mathematics. Additionally, datasets like the Knights and Knaves (K\&K) puzzles \cite{xie2024memorization} and LeetCode are commonly used for evaluating logical reasoning and coding proficiency.  

\noindent{\bf RewardBench}. The RewardBench \cite{lambert2024rewardbench} offers a comprehensive set of evaluations for reward models, encompassing key aspects such as chat, instruction-following, coding, safety, and other critical metrics for fine-tuned language models. The RwardBench dataset includes a combination of pre-existing evaluation prompt-completion pairs as well as newly curated examples specifically for this project.

\subsection{Numerical Results}
In this subsection, we will present some evaluations results of different alignment approaches. First, we will compare IRL approach \eqref{irl:algorithmic_pipeline} with other learning-from-demonstration approaches like SFT (Protocol \ref{prot:SFT}) and SPIN (Protocol \ref{prot:SPIN}) to illustrate their differences. Then we will compare a number of algorithms that leverage preference data to further improve the LLM alignment.  

\subsubsection{Demonstration-only setting}

We conduct experiments on the UltraChat dataset~\cite{ding2023enhancing}. Both the policy and reward models are initialized from the public checkpoint Mistral-7b-SFT-Beta\footnote{\texttt{HuggingFaceH4/mistral-7b-sft-beta, \url{https://huggingface.co/HuggingFaceH4/mistral-7b-sft-beta}}}, which is an SFT model fine-tuned on the UltraChat dataset using the base model Mistral-7B-v0.1\footnote{\texttt{mistralai/Mistral-7B-v0.1, \url{https://huggingface.co/mistralai/Mistral-7B-v0.1}}}.

Since UltraChat contains only demonstration data, we construct a synthetic preference dataset to train the reward model for IRL approaches. In each IRL iteration, we treat the demonstration data from UltraChat as preferred responses and the outputs generated by the IRL policy model as rejected responses. This method enables the creation of preference pairs without explicit human annotations. 

\begin{figure}[h]
    \subfigure[Policy Evaluations on Open LLM Leaderboard]{
        \includegraphics[width=0.47\linewidth]{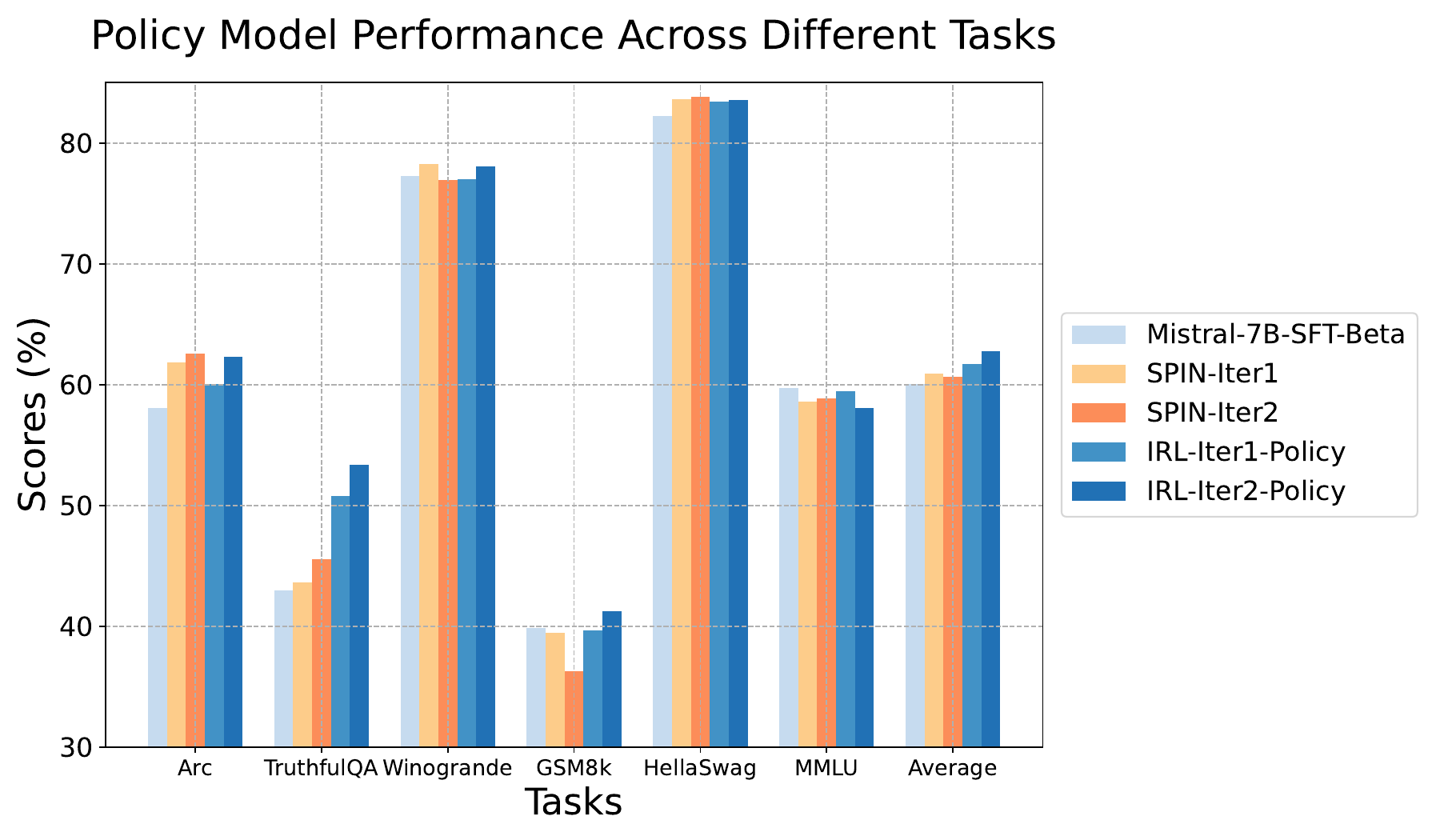}
        \label{subfig:policy_models}
    }
    \hfill
    \subfigure[Reward Evaluations on Reward Bench]{
        \includegraphics[width=0.47\linewidth]{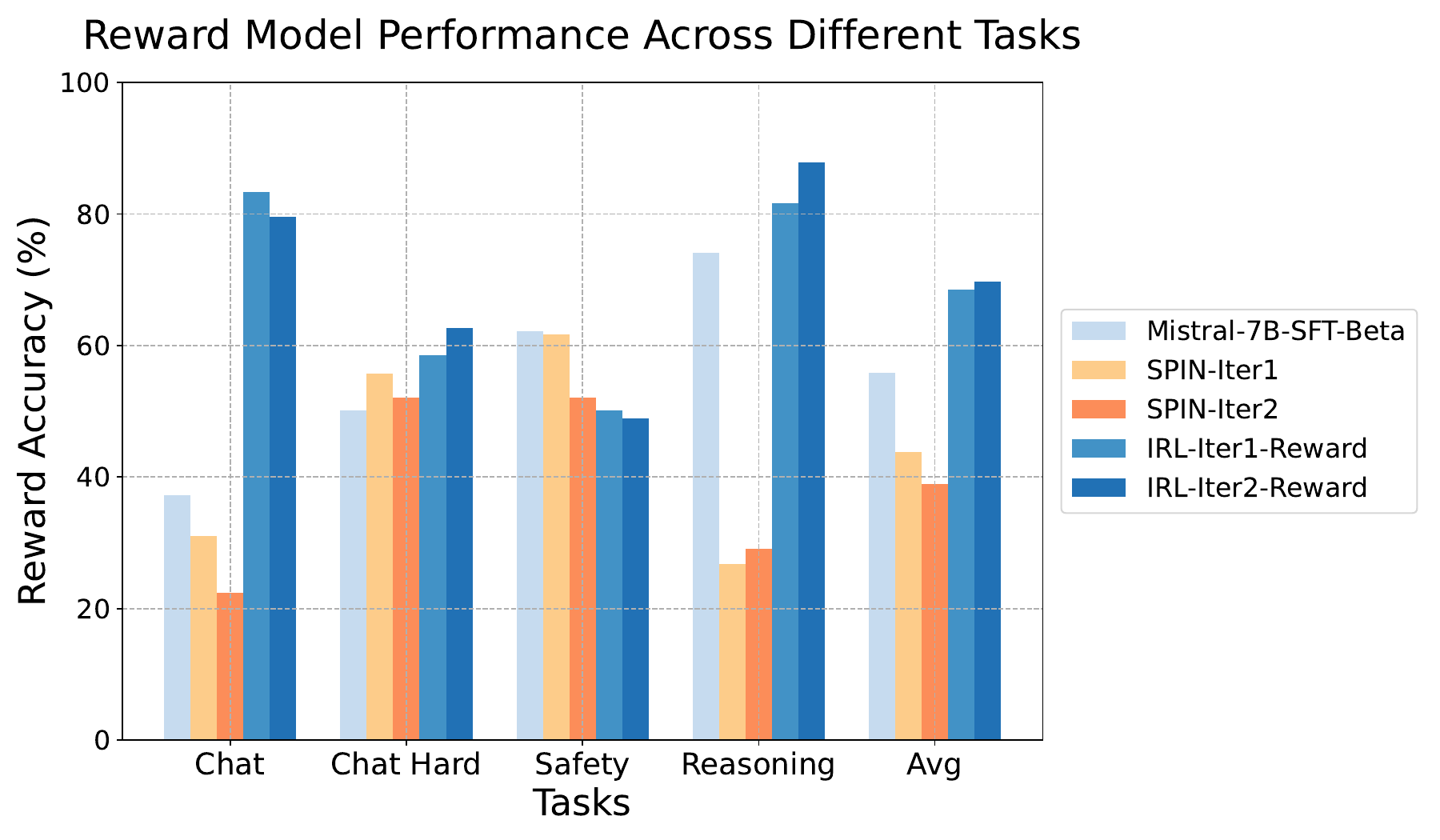}
        \label{subfig:reward_models}
    }
    \caption{{\textbf{Left:} Evaluation of policy models on the Open LLM Leaderboard. \textbf{Right:} Evaluation of reward models using RewardBench, where reward accuracy measures the proportion of preference pairs in which the reward model assigns a higher score to the preferred response.} Results show that high-quality demonstration datasets can effectively enhance both reward and policy models via the IRL method.}
    \label{figure:irl_evaluations}
\end{figure}

We assess the performance of reward models derived from several baselines using RewardBench\cite{lambert2024rewardbench}, focusing on categories pertinent to language understanding and generation. Since SPIN lacks an explicit reward model, we evaluate its performance based on an implicit reward formulated as $r(s,a) = \log \frac{\pi(a|s)}{\pi_{\rm ref}(a|s)}$, motivated by \eqref{eq:closed_form:reward_difference}. In our experiments, $\pi_{\rm ref}$ represents the SFT model. For the SFT model itself, we measure performance using the implicit reward $r(s,a) = \log \pi_{\rm SFT}(a|s)$.

The results, summarized in Table \ref{Table:MT-Bench}, indicate that the IRL-trained reward model outperforms both the base model (initialized from the SFT model) and the policy model trained with SPIN~\cite{chen2024self}. This demonstrates 
\begin{wraptable}{r}{9cm}
\footnotesize
    \begin{tabular}{c | c c c}
    \toprule
        Tasks & First turn & Second turn & Average \\
        \midrule
        \texttt{mistral-7b-sft-beta} & 5.66 & 5.09 & 5.37 \\
        \midrule
        \texttt{SPIN-Iter1} & 6.75 & 5.56 & 6.16 \\
        \texttt{SPIN-Iter2} & 3.18 & 3.41 & 3.29 \\
        \midrule
        \texttt{IRL-Iter1-Policy} & 6.71 & 5.96 & 6.33 \\
        \texttt{IRL-Iter2-Policy} & \textbf{7.01} & \textbf{6.19} & \textbf{6.60} \\
    \bottomrule
    \end{tabular}%
    \caption{\footnotesize Evaluation of Policy Models in MT-Bench.~{The results show that the IRL method can effectively improve the quality of policy models on multi-turn questions.}}
    \label{Table:MT-Bench}
\end{wraptable}
that integrating high-quality demonstration datasets with IRL methods, which generate synthetic preference pairs by combining demonstrations and model outputs, can significantly enhance reward model quality.
Furthermore, Figure \ref{fig:7B_downstream} illustrates the comparative evaluation of alignment methods when both demonstrations and preference dataset are available. The numerical results show that the IRL-based method is compatible with RLHF method to 
to enhance model alignment through integrating both demonstration and preference dataset.
\begin{figure}
\includegraphics[clip,width=0.95\linewidth]{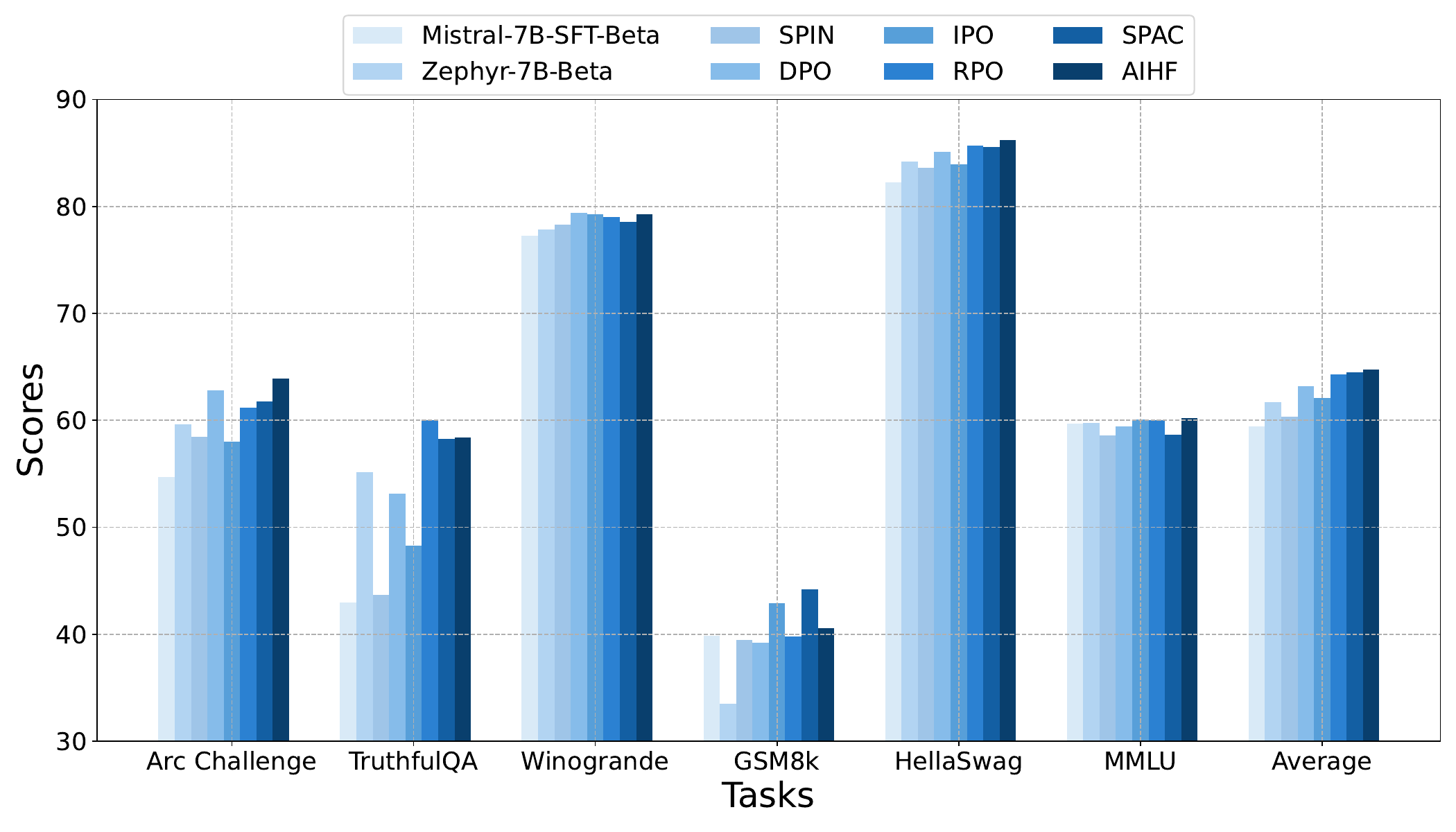}
\caption{Performance comparison of alignment methods. The AIHF-trained model demonstrates superior alignment across various downstream tasks.}
\label{fig:7B_downstream}
\vspace{-0.6cm}
\end{figure}

We evaluate various LLM policy models using the Open LLM Leaderboard~\cite{eval-harness} and MT-Bench~\cite{zheng2023judging}. As shown in Fig.\ref{figure:irl_evaluations} and Table\ref{Table:MT-Bench}, the IRL-based method consistently outperforms both the mistral-7b-sft-beta checkpoint and the SPIN method (regardless of whether self-generation is performed once or twice). These results demonstrate the effectiveness of leveraging high-quality demonstrations and synthetic preferences to improve language model performance in dialogue generation tasks.




\subsubsection{Aligning with Both Demonstrations and Pairwise Comparisons}
In this section, we evaluate several alignment algorithms, categorized as follows: {(1) Preference-only: DPO~\cite{rafailov2023direct}, IPO~\cite{azar2024general}, and SPAC~\cite{ji2024self}, which optimize policies solely based on preference data; (2) Demonstration-only: SPIN~\cite{chen2024self}; (3) Preference and demonstration: RPO~\cite{liu2024provably} and AIHF, which leverage both preference data and demonstrations.} For this experiment, we use UltraFeedback. 
as the preference dataset and UltraChat200k\footnote{Available at \url{https://huggingface.co/datasets/HuggingFaceH4/ultrachat_200k}.} as the demonstration dataset. The base model for our experiments is mistral-7b-sft-beta\footnote{Available at \url{https://huggingface.co/HuggingFaceH4/mistral-7b-sft-beta}.} \cite{jiang2023mistral}, which is also used to initialize the reward model.


\begin{table}[t]
    \centering
    \footnotesize
    \begin{tabular}{c |c c c c c }
        \toprule
        \textbf{Reward Model} & \textbf{Chat} & \textbf{Chat Hard} & \textbf{Safety} & \textbf{Reasoning} & \textbf{Average} \\ \midrule
        DPO Reward Model & 37.43\% & 55.92\% & 64.14\% & 47.33\% & 51.21\% \\ 
        BTL Reward Model  & \textbf{95.11\%} & \textbf{56.58\%} & 63.69\% & 69.22\% & 71.15\% \\
        AIHF Reward Model &  94.41\%& 55.37\% & \textbf{63.98\%} & \textbf{76.75\%} & \textbf{72.63\%} \\\bottomrule
    \end{tabular}
    \caption{Evaluation of reward models in Reward-Bench. It can be seen that the reward model estimated by AIHF which incorporates {\it both} demonstrations  and preference achieves better performance.}\label{tab:reward_bench}
    \vspace{-0.5cm}
\end{table}

Fig. \ref{fig:7B_downstream} evaluates the quality of the aligned models trained on different methods. Furthermore, in Table \ref{tab:reward_bench}, we also evaluate the reward models estimated using different methods (DPO, standard preference learning and AIHF) over the popular benchmark RewardBench \cite{lambert2024rewardbench}. {The BTL reward model is trained in the RLHF pipeline using the loss function defined in \eqref{eq:RLHF_loss}}. The DPO reward is obtained by training a DPO loss on preference data (UltraFeedback) and use the following implicit reward  $r(s,a) = \log \frac{\pi(a|s)}{\pi_{\rm ref}(a|s)}$ where $\pi_{\rm ref}$ is the SFT model. For AIHF, we use the demonstration data UltraChat and generate the non-preferred sample using the base model (mistral-7b-stf-full), then train a reward with UltraFeedback and UltraChat data combined. It can be seen that the reward model estimated by AIHF which incorporates {\it both} demonstrations  and preference achieves better performance.


\section{Connection of Alignment to Signal Processing}

Before we conclude this article, we would like to point out that the SP community has a wealth of expertise that can contribute significantly to alignment research. Given the complexity of aligning LLMs with human preferences, SP methodologies offer valuable tools and frameworks that can improve both the theoretical and practical aspects of alignment. 

The connection between large-scale model alignment and SP can be made precise once we recognize that the core challenge of alignment is \emph{inference from indirect, noisy, and often incomplete measurements of human intent}. Many mathematical tools originally developed in SP, such as inverse problems, denoising and robust estimation all provide rigorous frameworks for tackling precisely these challenges. Below, we elaborate on a few of these connections.

First, alignment can be viewed as a broad class of \emph{inverse problems}, in which both the human intent and the model’s reasoning trajectory act as latent variables that must be inferred from indirect and noisy observations. In this view, human feedback (e.g., pairwise preferences or rankings) constitutes a noisy measurement of an underlying latent \emph{reward signal} that reflects human values, while the model’s internal reasoning process represents a second latent structure that mediates how these preferences are expressed in behavior. This setting is analogous to classical SP problems such as image deblurring or tomography, where one seeks to recover a clean signal $x$ from degraded observations $y = A x + \varepsilon$. Here, both the human reward signal and the model’s internal reasoning can be regarded as hidden components of $x$, and the alignment algorithm attempts to reconstruct them from limited feedback data $y$.

Recent work has begun to explicitly exploit this latent-variable perspective. For example, \cite{zhongbrite} formulates reasoning-based alignment as a joint inference problem: it treats an LLM’s internal rationales as hidden variables and alternates between (i) inferring plausible reasoning chains using reward-guided sampling, and (ii) updating model parameters to maximize the joint log-likelihood of the latent rationales and final outputs. This two-stage optimization is formally equivalent to an \emph{Expectation--Maximization (EM)} procedure widely used in SP and statistics to solve inverse problems with hidden variables. In this context, the E-step corresponds to estimating latent model reasoning signals, while the M-step updates model parameters given these inferred signals. Such a formulation provides a principled link between SP-style latent-variable estimation and alignment objectives, and clarifies how  human preferences or reasoning traces can be reconstructed and aligned through iterative inference.

Such latent variable perspectives also extend to the IRL problem discussed earlier, which can be viewed as solving inverse problems over high-dimensional function spaces. Classical SP regularization techniques such as Tikhonov penalties, total-variation priors, or sparsity constraints can provide principled ways to stabilize these ill-posed estimation problems, particularly when human feedback is sparse or inconsistent.

Moreover, as alignment research expands to multi-modal domains involving text, images, speech, and video, SP expertise in multi-modal signal fusion, feature extraction, and cross-modal learning becomes increasingly relevant. The SP community’s long-standing experience in designing architectures that integrate heterogeneous signal sources can directly inform the development of more effective and interpretable multi-modal alignment mechanisms.





\section{Concluding Remarks}\label{sec:conclusion}

In this work, we provided a tutorial on aligning LLMs with human feedback, exploring both mathematical foundations and state-of-the-art algorithmic approaches. 
Despite significant progress in aligning LLMs with human values and preferences, several open challenges remain.

 LLM alignment methods continue to face challenges related to scalability and effectiveness. Learning robust and high-quality policies from reward models is challenging and requires extensive human annotation and computation. Existing RL-based fine-tuning methods, such as PPO, demand careful hyperparameter tuning and require a huge amount of compute resources. Moreover, reward hacking remains a significant problem -- where models find unintended shortcuts to maximize reward scores without genuinely improving response alignment. Instead of producing responses that align with human intent, models can generate outputs optimized purely for high reward function scores, potentially leading to unreliable behavior. Addressing these concerns requires the development of more efficient RL algorithms, robust and diverse reward functions (e.g., a combination of trained reward functions and the verifiable ones mentioned in Sec. \ref{sub:verifier}), and hybrid approaches that integrate supervised learning with RL to improve scalability and alignment fidelity.

 \section*{Acknowledgments}
 Luca Viano was supported by the Swiss Data Science Center under fellowship number P22\_03. This work was supported as part of the Swiss AI Initiative, through a grant from the ETH Domain and
computational resources provided by the Swiss National Supercomputing Centre (CSCS) under the
Alps infrastructure. Research was sponsored by the Army Research Office and was accomplished
under Grant Number W911NF-24-1-0048. This work was funded by the Swiss National Science
Foundation (SNSF) under grant number 2000-1-240094.



\maketitle

\bibliographystyle{IEEEbib}
\bibliography{Refs/refs,Refs/reference,Refs/iclr2025_conference}

\end{document}